\documentclass[10pt,twocolumn,letterpaper]{article}

\usepackage{cvpr}              % To produce the CAMERA-READY version

\usepackage{graphicx}
\usepackage{amsmath}
\usepackage{amssymb}
\usepackage{booktabs}
\usepackage{comment}

\usepackage[
    style=ieee,
    sorting=nyt,
    citestyle=numeric-comp,
    giveninits=false,
    dashed=false,
    maxnames=4,
    minnames=4]{biblatex}
\AtEveryBibitem{%
  \clearlist{publisher} % Remove doi
  \clearfield{issn} % Remove issn
  \clearfield{doi} % Remove doi
  \clearfield{isbn} % Remove doi
  \clearfield{pages} % Remove doi
  \clearfield{volume} % Remove doi
  \clearfield{month} % Remove doi
  \clearfield{number} % Remove doi

  \ifentrytype{online}{}{% Remove url except for @online
    \clearfield{url}
  }
}

\DeclareSourcemap{ 
    \maps[datatype=bibtex]{
      \map{
           \step[fieldsource=journal, match=\regexp{International}, replace=\regexp{Int.}]
           \step[fieldsource=booktitle, match=\regexp{International}, replace=\regexp{Int.}]
           \step[fieldsource=journal, match=\regexp{Conference}, replace=\regexp{Conf.}]
           \step[fieldsource=booktitle, match=\regexp{Conference}, replace=\regexp{Conf.}]
           \step[fieldsource=journal, match=\regexp{Transactions}, replace=\regexp{Trans.}]
           \step[fieldsource=booktitle, match=\regexp{Transactions}, replace=\regexp{Trans.}]
          }
    }      
}
\AtBeginBibliography{\footnotesize}
\usepackage[breaklinks,colorlinks]{hyperref}

\usepackage[capitalize]{cleveref}
\crefname{section}{Sec.}{Secs.}
\Crefname{section}{Section}{Sections}
\Crefname{table}{Table}{Tables}
\crefname{table}{Tab.}{Tabs.}

\newcommand{\algname}{\mbox{Motron}}
\newcommand{\emphalgname}{\emph{\algname}}

\newcommand{\rom}[1]{\uppercase\expandafter{\romannumeral #1\relax}}

\newcommand\rebuttal[1]{\textcolor{blue}{#1}}
\renewcommand\rebuttal[1]{#1}
\newcommand\rebuttalrm[1]{}

\def\cvprPaperID{11089} % *** Enter the CVPR Paper ID here
\def\confName{CVPR}
\def\confYear{2022}

\begin{document}

\title{\algname: Multimodal Probabilistic Human Motion Forecasting}

\author{Tim Salzmann${}^1$, Marco Pavone${}^{2, 3}$ and Markus Ryll${}^1$\\
${}^1$ Technical University of Munich \hspace{1em} ${}^2$ Stanford University \hspace{1em} ${}^3$ NVIDIA Research\\
{\tt\small \{tim.salzmann, markus.ryll\}@tum.de} \hspace{1em} {\tt\small pavone@stanford.edu}

}
\maketitle

%%%%%%%%% ABSTRACT
\begin{abstract}
Autonomous systems and humans are increasingly sharing the same space. Robots work side by side or even hand in hand with humans to balance each other's limitations. 
%Autonomous vehicles share the road with human operated cars, cyclists and pedestrians. 
Such cooperative interactions are ever more sophisticated. Thus, the ability to reason not just about a human's center of gravity position, but also its granular motion is an important prerequisite for human-robot interaction. 
Though, many algorithms ignore the multimodal nature of humans or neglect uncertainty in their motion forecasts.
We present \algname, a multimodal, probabilistic, graph-structured model, that captures human's multimodality using probabilistic methods while being able to output deterministic \rebuttal{maximum-likelihood} motions and corresponding confidence values for each mode.
Our model aims to be tightly integrated with the robotic planning-control-interaction loop; outputting physically feasible human motions and being computationally efficient. % to run on embedded devices in real-time.
%Our model is designed to be tightly integrated with robotic planning and control frameworks; it is capable of producing predictions that are conditioned on ego-agent motion plans.
We demonstrate the performance of our model on several challenging real-world motion forecasting datasets, outperforming a wide array of generative\rebuttal{/variational} methods while providing state-of-the-art \rebuttal{single-output} motions if required. Both using significantly less computational power than state-of-the art algorithms.
%providing state-of-the-art deterministic capabilities.
\end{abstract}

%%%%%%%%% BODY TEXT
\vspace{-1.5em}
\section{Introduction}\label{sec:Intro}
The key desideratum of autonomous systems is to provide added-value to humans while ensuring safety. Traditionally, safety aspects limit such robots to low-risk tasks with minimal human interaction. An understanding of humans and their distribution of feasible anticipated movement is key to develop safe, risk-aware human-interactive autonomous systems. Such systems could operate in closer proximity to humans, performing tasks involving higher levels of interaction, providing enhanced added value.
%\cref{fig:hero} illustrates a handover scenario where predicting the human's motion is key to the autonomous system's task and trajectory planning.
% \begin{figure}[t]
%     \centering
%     \includegraphics[width=.75\linewidth]{figures/hero_figure.png}
%     \caption{Exemplary scenario of a drone handing over an object to a human. Predicting the human's motion is key to a successful interaction during all stages of the maneuver.%: Approach, Handover, and Departure.
%     }
%     \label{fig:hero}
% \end{figure}
\begin{figure}
    \begin{subfigure}{0.49\linewidth}
        \frame{\includegraphics[trim=2.1cm 0.8cm 2.1cm 2.4cm, clip,width=\linewidth]{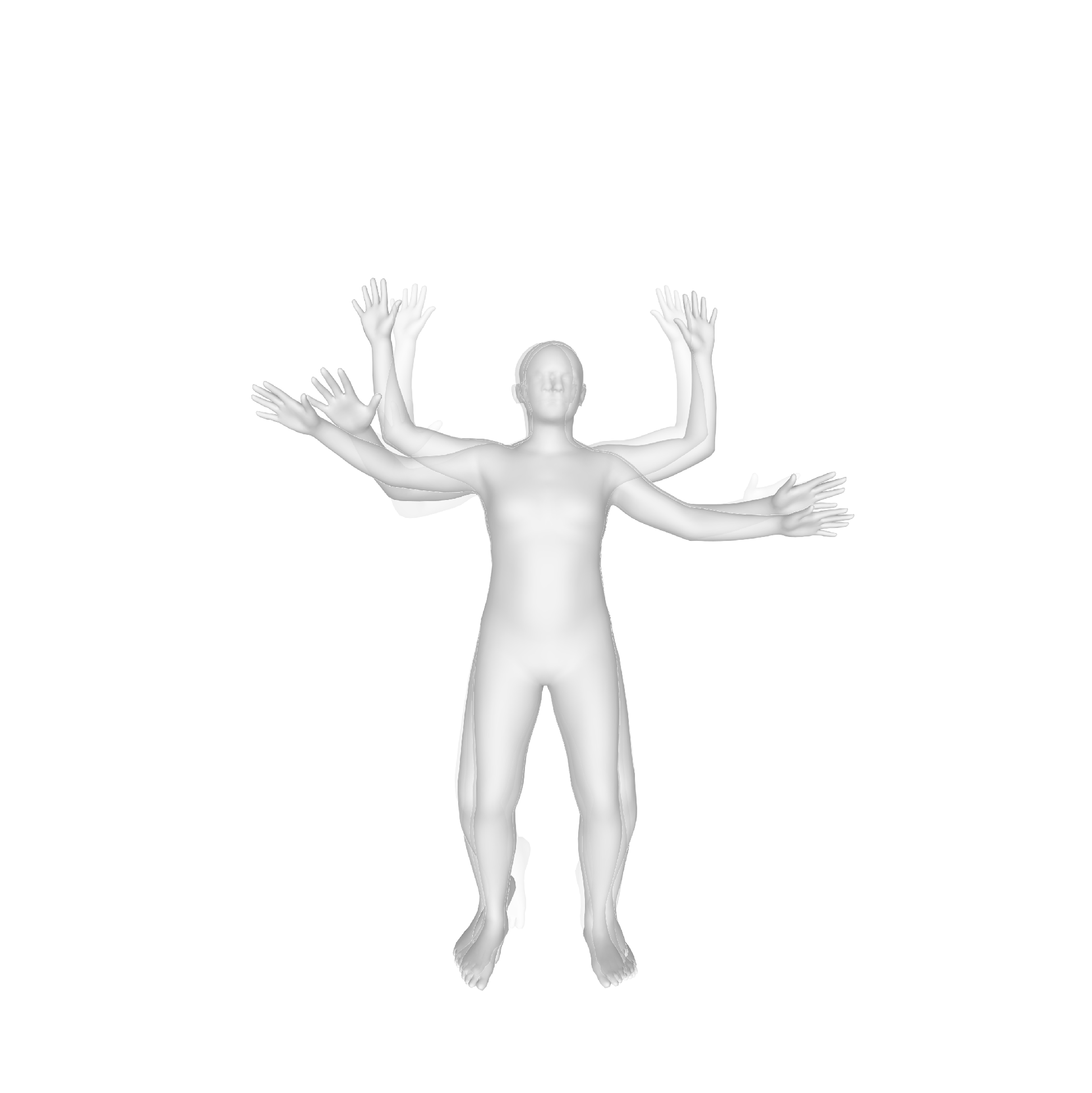}}
    \end{subfigure}
    \begin{subfigure}{0.49\linewidth}
        \frame{\includegraphics[trim=2.1cm 0.8cm 2.1cm 2.4cm, clip,width=\linewidth]{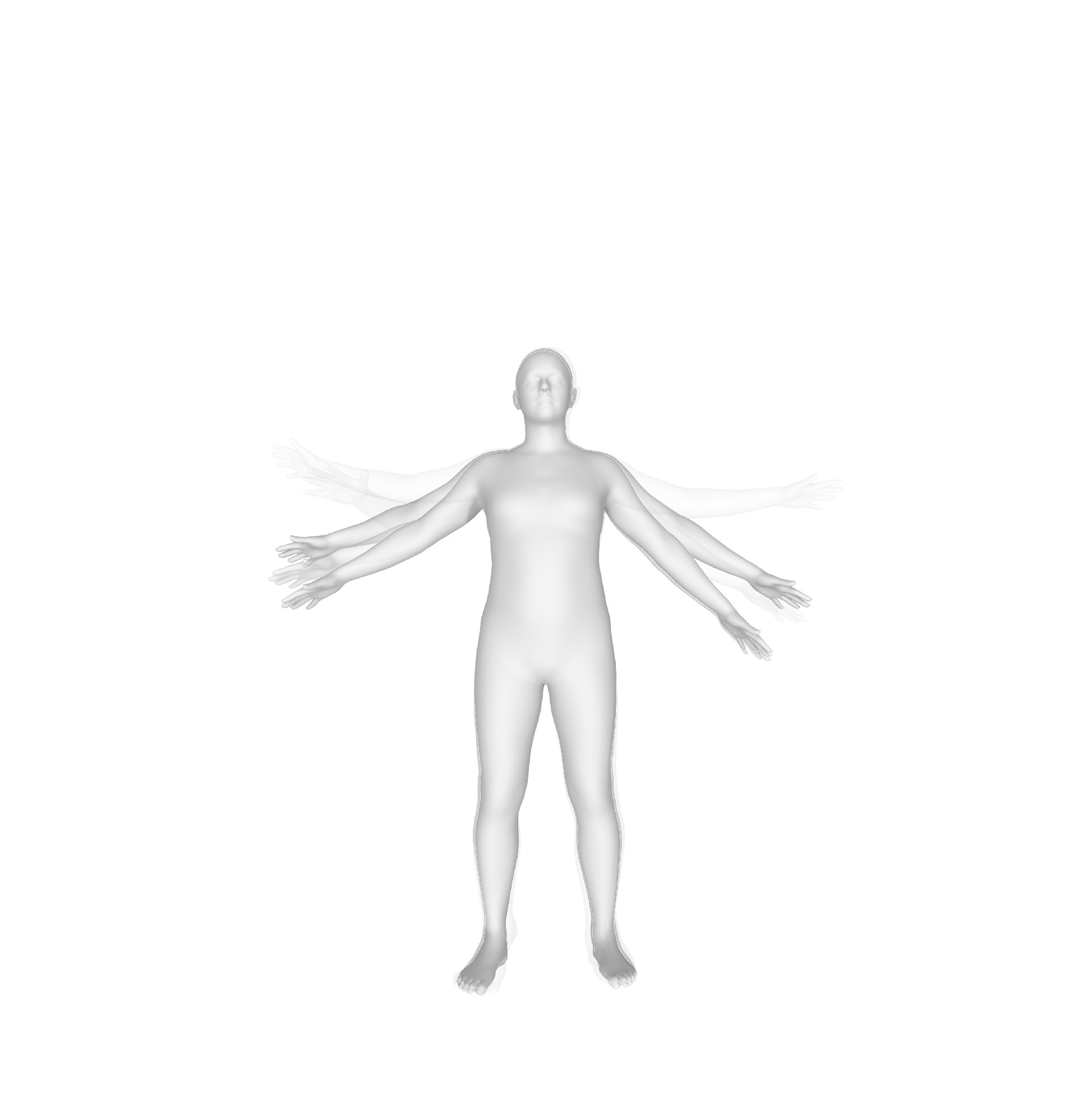}}
    \end{subfigure}%
    \caption{\rebuttal{Two examples of multiple possible poses at a given prediction time weighted by their probability. Left: Start of high variance jumping motion. Right: Landing from jump. Lower body is predicted with low uncertainty as feet will come to a stop during landing. Our parametric output distribution captures the full multimodal uncertainty in human motion; enabling subsequent evaluation of samples, individual modes, or most-likely motion as well as their respective confidence values.}}
    \label{fig:hero}
\end{figure}

However, capturing the complexity of human motion in a computational model is challenging due to the multitude of continuous movement possibilities (multimodality), even within fixed boundaries of physical limitations. Traditionally, over-conservative systems rely solely on those constraints to ensure safety, while\rebuttalrm{ deterministic} predictive \rebuttal{single-motion-output} methods discard the potential of many high-level futures in favor of a single possible motion.

In contrast to such deterministic regressors,\rebuttalrm{generative} \rebuttal{Monte-Carlo method} models aim to produce samples of human future motion. However, for different reasons (e.g., simulation purposes), they prioritize generative diversity over representing actual plausible motions. For robotic use cases, we aim for samples to represent the underlying distribution of possible motions; or even better a parametric description thereof (\cref{fig:hero}).

Both,\rebuttalrm{deterministic and generative} \rebuttal{single-output and sampled}, human motion predictions have been developed independently to maximize their strengths in accuracy and diversity. Still, many of them have been developed without directly accounting for real-world robotic use cases; they settle for diversity instead of accuracy, ignore physical boundaries, and are  computationally too expensive.
We target robotic use cases, where predictions are used in a control loop. As such, a model has to capture the underlying distribution of possible human motions without being over-confident or over-diverse within an imminent time horizon; \emph{combining} and \emph{balancing} the desiderata of both research strands.
In consequence, we are interested in developing a human motion prediction model that
(\rom{1})~represents the inherent multimodal structure of human motion;
(\rom{2})~accurately captures humans' probabilistic and diverse nature while obeying \rebuttal{rigid skeleton constraints};
(\rom{3})~by directly incorporating structural information of the human skeleton;
(\rom{4})~while being able to deal with imperfect data;
(\rom{5})~outputting the maximum amount of information for the use of subsequent systems.

%In this work we present \emphalgname{}, an open and extensible approach which produces physically-feasible motion forecasts capturing structural, multimodal, and probabilistic human characteristics.

%\tstodo{Do I need another sentence here?}

To achieve those desiderata, our contribution is threefold: 
First, we describe a new efficient way to model graph-structured problems where nodes have a fixed semantic class, usable for generic graph-structured problems. 
Secondly, we present \emphalgname, which uniquely uses a probabilistic output structure based on the Concentrated Gaussian distribution in SO3 and a parallel weight sharing approach incorporating the skeleton's structure. It is designed to mirror the multimodal and uncertain nature of humans. %To the best of our knowledge, this is the first algorithm capable of producing a parametric distribution over future human skeleton configurations. 
\emphalgname's flexible output structure is designed to serve downstream robotic modules such as motion planning, decision making, and control.
Finally, we evaluate our model on the fulfillment of our desiderata using\rebuttalrm{deterministic and generative} \rebuttal{single-output} metrics \rebuttal{\textit{and} metrics based on samples}; combining both research strands. We outperform an extensive selection of\rebuttalrm{generative} \rebuttal{Monte-Carlo based} motion prediction methods while showing state-of-the-art performance on\rebuttalrm{deterministic} \rebuttal{single-output} evaluation procedures using a variety of metrics and datasets. Our contributions are substantiated by a thorough ablation study. We further show that \emphalgname{} can deal with occluded data often present in real-world applications by including the additional uncertainty in its output.

\vspace{-1em}
\section{Related Work}
{\bf Human Motion Forecasting.} The release of the H3.6M dataset \cite{Ionescu2014Human3.6M:Environments} in 2014 stimulated a wealth of research in the field of human motion prediction. While early works were based on traditional approaches such as Markov Models \cite{Pavlovic2000LearningMotion, Lehrmann2014EfficientMotion} or Gaussian Processes \cite{Wang2008GaussianMotion}  the advances of deep learning has taken over recent approaches.
Algorithms now are largely based on\rebuttalrm{deterministic} \rebuttal{single-motion-output} regressors \cite{Martinez2017OnNetworks, Fragkiadaki2015RecurrentDynamics, Ghosh2017LearningPredictions, Pavllo2019ModelingNetworks}; improving performance by incorporating human skeleton structural information into their architectures  \cite{Jain2016Structural-RNN:Graphs, Li2020DynamicPrediction}. Li \etal \cite{Li2020DynamicPrediction} merge nodes to create meta graphs of different scale to extract higher level information. Two concepts of capturing temporal influences have emerged: Recurrent Neural Networks (RNNs) and transforming time-series to a frequency domain \cite{Aksan2021APrediction, Mao2020HistoryAttention, Zhang2021WeMove}, where the latter is not agnostic to varying history horizons. Recently, Transformer \cite{Vaswani2017AttentionNeed} architectures have been explored overcoming RNNs shortcomings in capturing longer time series \cite{Mao2020HistoryAttention, Tang2018Long-TermDynamics}.
While generative and \rebuttal{variational} approaches have emerged as state-of-the-art in trajectory forecasting \cite{Salzmann2020Trajectron, Tang2019MultiplePrediction, Kosaraju2019Social-BiGAT:Networks} for their plethora of captured information, they largely exist in parallel to research on\rebuttalrm{deterministic} \rebuttal{single-output} regressors in human motion forecasting. The field is split in approaches using Generative Adversarial Networks (GANs) \cite{Gui2018AdversarialPrediction, Barsoum2018HP-GAN:GAN, Kundu2018BiHMP-GAN:GAN, Gurumurthy2017DeLiGAN:Data} and (Conditional) Variational Autoencoders ((C)VAEs) \cite{Yuan2020DLow:Prediction, Zhang2021WeMove, Yuan2020DiverseProcesses}. Of these, similar to the field of trajectory prediction, adapted versions of the VAE frameworks show better results \cite{Yuan2020DLow:Prediction}. 

Surprisingly, these two fields of\rebuttalrm{deterministic and generative} \rebuttal{single-output and probabilistic} motion prediction are highly disentangled, following their respective distinct experimentation protocols. In\rebuttalrm{deterministic} \rebuttal{single-motion-output} setups, it is, for example, common to predict one future second. In contrast, previous\rebuttalrm{generative} \rebuttal{probabilistic} works set their focus on producing plausible diverse motion samples over a longer prediction horizon (2s); at which point the true human motion is fraught with uncertainty.

\rebuttal{By slight abuse of terminology, we will distinguish the two common types of motion prediction algorithms in the remainder of this work. Monte-Carlo based models which can produce samples from a distribution of future motions will be referenced as \textit{probabilistic} models while single-motion-output models will be referenced as \textit{deterministic} models for their lack of uncertainty awareness.}

\begin{comment}
\begin{table}[t]
\fontsize{8}{8}\selectfont
\centering
\caption{A summary of recent state-of-the-art human motion forecasting methods and their addressed desiderata.}
\begin{tabular}{l|cccccc}
\toprule
\textbf{Method} & G & P & C & DTH & CF & OS \\ \midrule
AGED \cite{Gui2018AdversarialPrediction} &  & & \checkmark & \checkmark & \\
DMGMM \cite{Li2020DynamicPrediction}&  &  & \checkmark & \checkmark & & \checkmark\footnotemark \\
DLow \cite{Yuan2020DLow:Prediction} & \checkmark & & & \checkmark & \checkmark & \checkmark\footnotemark\\
LCP-VAE \cite{Aliakbarian2021ContextuallyPrediction} & \checkmark & & & \checkmark & \checkmark & ?\footnotemark \\
HistRepItself \cite{Mao2020HistoryAttention}  & & & \checkmark & & \checkmark & \checkmark\footnotemark\\
\midrule 
Our Work & \checkmark & \checkmark & \checkmark & \checkmark & \checkmark & \checkmark\\
\bottomrule
\end{tabular}

Legend: G = Generative, P = Probabilistic Output, C = Adheres to rigid skeleton constraints, DTH = Dynamic Time Horizon, CF = Computationally Feasible, OS = Open Source

\label{tab:related_work}
\end{table}
\addtocounter{footnote}{-3}
\stepcounter{footnote}\footnotetext{\url{https://github.com/limaosen0/DMGNN}}
\stepcounter{footnote}\footnotetext{\url{https://github.com/Khrylx/DLow}}
\stepcounter{footnote}\footnotetext{Not openly accessible at time of submission.}
\stepcounter{footnote}\footnotetext{\url{https://github.com/wei-mao-2019/HisRepItself}}
\end{comment}

{\bf Directional Probabilistic Learning.} 
In robotic applications, such as filtering, the use of directional statistics for rotational systems has been proven useful \cite{Glover2014, Kurz2014EfficientDistribution, Kurz2017StatisticsLibDirectional, Gilitschenski2014EfficientApproximations}. The most common distributions for modeling rotational uncertainty are the Bingham \cite{Bingham1974AnSphere}, the von Mises–Fisher \cite{Fisher1953DispersionSphere}, the Projected Gaussian \cite{LangApproximationQuaternions}, and using a Concentrated Gaussian in SO(3)\cite[Chapter~7.3.1]{Barfoot2021StateRobotics}. Advances in deep probabilistic learning, however, are mainly focused on learning distributions in vector space \cite{Bishop1994MixtureNetworks, Bingham2018Pyro:Programming, Salzmann2020Trajectron}. Thus, of the rotational distributions, only the Bingham distribution has been applied to probabilistic deep learning \cite{Gilitschenski2020DeepLoss, Peretroukhin2020AUncertainty}. In \cite{Gilitschenski2020DeepLoss}, Gilitschenski \etal directly learn the parameters of a Bingham distribution representing the orientation of an object in an image.
%To the best of our knowledge, there is no approach leveraging probabilistic learning within the field of human motion prediction.

{\bf Graph Neural Networks.} 
Besides other concepts like message passing, convolution, and aggregation \cite{Zhou2020GraphApplications} the concept of attention, first introduced for temporal dependencies \cite{Vaswani2017AttentionNeed}, has been applied to graph-structured problems by Veličković \etal \cite{Velickovic2017GraphNetworks} as Graph Attention Networks (GAT). Recently those GATs have been included in temporal networks such as LSTMs by replacing the linear transformations in each RNN cell with a Graph Attention Layer \cite{Wu2018GraphForecasting}. Within human motion forecasting Graph Convolution Networks (GCNs) \cite{Mao2019LearningPrediction, Mao2020HistoryAttention} have shown good performance. Salzmann \etal \cite{Salzmann2020Trajectron} model dependencies between different entities for trajectory prediction using a sequential message passing algorithm. This, however, becomes computational infeasible for larger number of node types.

%\cref{tab:related_work} provides a detailed breakdown of recent state-of-the-art approaches and their consideration of desiderata.

\vspace{-1em}
\section{Problem Formulation}\label{sec:problem_formulation}
We aim to generate plausible motion distributions for a fixed number $N$ of human skeleton nodes (joints) $n_1,...,n_{N}$. Each node $n_i$ is assigned to a semantic class $S_i$, e.g. Elbow, Knee, or Hip.
%Nodes are structured in a static tree where each node can have multiple children and a single parent. $A(n_i)$ defines the list of all ancestor nodes of $n_i$.
At time $t$, given the $D$-dimensional state $\mathbf{s} \in \mathbb{R}^D$ of each node
%$\mathbf{x}_{i}^{(t)} \in \mathbb{R}^{D},\ i \in \{1, \dots, N(t)\}$,
and all of their histories for the previous $H$ timesteps, which we denote as $\mathbf{x} = \mathbf{s}_{1,\dots,N}^{(t - H : t)} \in \mathbb{R}^{(H + 1) \times N \times D}$,
%the state of each node $\mathbf{x}_{i}^{(t)} \in \mathbb{R}^{D},\ i \in \{1, \dots, N\}$, all of their histories for the previous $H$ timesteps $\mathbf{x}_{1,\dots,N}^{(t - H : t-1)} \in \mathbb{R}^{H \times N \times D}$
%, and additional information available to each node $I_{1,\dots,N}^{(t)}$,
we seek a distribution over all nodes' future states for the next $T$ timesteps $\mathbf{y} = \mathbf{s}_{1,\dots,N}^{(t + 1 : t + T)} \in \mathbb{R}^{T \times N \times D}$, which we denote as $p(\mathbf{y} \mid \mathbf{x})$. 
%Note that the number of past and future timesteps are variable, \ie $H = f(n_i, S_i, t)$ and $T = g(n_i, S_i, t)$, to incorporate limited history due to sensor range and variable foresight requirements depending on the situation.

%One of our key desiderata is the ability to produce trajectories that take into account ego-agent motion plans, for downstream use in motion planning, decision making, and control. Thus, we also assume that we know the ego-agent's future motion plan for the next $T$ timesteps, $\mathbf{y}_R^{(t + 1 : t + T)}$, a quantity that is readily available online from previous motion plans or current motion hypotheses, \eg if using \emphalgname{} to evaluate a set of possible motion plans.

\section{Preliminaries}\label{sec:prob_rot}
{\bf Quaternion Representation.}
The rotational nature of the human anatomy presents a challenge to neural networks. Commonly used rotation representations in $\mathbb{R}^3$, Euler angles and exponential maps, suffer from singularities, discontinuity, and non-uniqueness \cite{Pavllo2019ModelingNetworks, Grassia1998PracticalMap}; all properties contrary to neural network characteristics. Outside of deep learning, however, quaternions have long been established as the default rotational representation. As a consequence of their properties, interpretability, and common use in robotics we choose quaternions as the data representation throughout our model. Thus the input state $\mathbf{x}_{i}^{(t)} = [\mathbf{q}_{i}^{(t)}, \dot{\mathbf{q}}_{i}^{(t)}]$ is defined as the concatenation of the rotation in quaternion representation and its time differential, where $\mathbf{q}_{i}^{(t)} = \mathbf{q}_{i}^{(t-1)} \tiny{\odot} \dot{\mathbf{q}}_{i}^{(t)}$; resulting in a $D=8$ dimensional state. 

%Consequently, we seek the models output to be a distribution over quaternions $\mathbf{y}_{1,\dots,N(t)}^{(t + 1 : t + T)} = q_{1,\dots,N(t)}^{(t + 1 : t + T)}$.

{\bf Probabilistic Rotations.}
We use the Concentrated Gaussian distribution $\mathcal{N}_{SO(3)}$ \cite[Chapter~7.3.1]{Barfoot2021StateRobotics} to model a probability distribution over the rotation group SO(3). A probabilistic rotation is given as
\begingroup
\setlength\abovedisplayskip{4pt}
\setlength\belowdisplayskip{4pt}
\begin{equation}
    \mathbf{R} = \text{exp}(\mathbf{\epsilon}\hat{~})%_{q}
    \bar{\mathbf{R}}
\end{equation}
\endgroup
where $\bar{\mathbf{R}}$ is a 'large', noise-free, nominal rotation, exp is the exponential map, $\hat{~}$ is a linear, skew-symmetric Lie algebra operator, %$(\cdot)_q$ transforms a rotation matrix into its equivalent quaternion 
and $\mathbf{\epsilon} \in \mathbb{R}^3$ is a 'small', noisy component:
\begingroup
\setlength\abovedisplayskip{4pt}
\setlength\belowdisplayskip{4pt}
\begin{equation}
    \mathbf{\epsilon} \sim \mathcal{N}(\mathbf{0}, \mathbf{\Sigma})
\end{equation}
\endgroup
The distribution's \textit{p.d.f} is defined as 
\begingroup
\setlength\abovedisplayskip{4pt}
\setlength\belowdisplayskip{4pt}
\begin{equation}
    p(\mathbf{R} \mid \bar{\mathbf{R}}, \mathbf{\Sigma}) = %\frac{1}{\sqrt{(2\pi)^3 \text{det}(\mathbf{\Sigma})}}
    \frac{1}{Z}
    \text{e}^{-\frac{1}{2} (\text{ln}(\mathbf{R}\bar{\mathbf{R}}^T)\check{~})^{T}\mathbf{\Sigma}^{-1} (\text{ln}(\mathbf{R}\bar{\mathbf{R}}^T)\check{~})}
\end{equation}
\endgroup
where $Z$ is the Gaussian normalization constant, ln is the inverse of the exponential map, and $\check{~}$ is the inverse linear, skew-symmetric Lie algebra operator.

Unlike the Bingham or Von Mises-Fisher, the Concentrated Gaussian distribution supports the analytical composition of rotations which is a necessary property as we use differential quaternions as intermediate output representation \rebuttal{to leverage a residual connection concept} (this is justified in \cref{sec:ablation}). A probabilistic multiplication of two rotations $\mathbf{R}_3 = \mathbf{R}_1 \mathbf{R}_2$ is expressed as
\begingroup
\setlength\abovedisplayskip{4pt}
\setlength\belowdisplayskip{4pt}
\begin{align}
\begin{split}
    \mathbf{R}_3    = \mathbf{R}_1 \mathbf{R}_2 
                    = \text{exp}(\mathbf{\epsilon}_1\hat{~}) \bar{\mathbf{R}}_1 \text{exp}(\mathbf{\epsilon}_2\hat{~}) \bar{\mathbf{R}}_2
\end{split}
\end{align}
\endgroup
without approximation we have
\begingroup
\setlength\abovedisplayskip{4pt}
\setlength\belowdisplayskip{4pt}
\begin{align}
\begin{split}
    \mathbf{R}_3    &= \text{exp}(\mathbf{\epsilon}_1\hat{~}) \text{exp}((\bar{\mathbf{R}}_1\mathbf{\epsilon}_2)\hat{~}) \bar{\mathbf{R}}_1 \bar{\mathbf{R}}_2 \\
                    &= \text{exp}(\mathbf{\epsilon}_1\hat{~}) \text{exp}(\mathbf{\epsilon}_{12}\hat{~}) \bar{\mathbf{R}}_3
\end{split}
\end{align}
\endgroup
using first order Baker-Campbell-Hausdorff approximation we get
\begingroup
\setlength\abovedisplayskip{4pt}
\setlength\belowdisplayskip{4pt}
\begin{align}\label{eq:comp}
    \mathbf{R}_3    &= \text{exp}((\mathbf{\epsilon}_1 + \mathbf{\epsilon}_{12})\hat{~}) \bar{\mathbf{R}}_3 \\
    \mathbf{R}_3 &\sim \mathcal{N}_{SO(3)}(\bar{\mathbf{R}}_1 \bar{\mathbf{R}}_2, \mathbf{\Sigma}_1 + \bar{\mathbf{R}}_1 \mathbf{\Sigma}_2 \bar{\mathbf{R}}_1^T)
\end{align}
\endgroup
To represent multimodality, we define our output structure as a Concentrated Gaussian Mixture Model in SO(3) ($\mathcal{N}^{\pi}_{SO(3)}$). For $q \sim \mathcal{N}^{\pi}_{SO(3)}(\pi_i, \bar{\mathbf{R}}_i, \mathbf{\Sigma}_i)$, the \textit{p.d.f} is given as
\begingroup
\setlength\abovedisplayskip{4pt}
\setlength\belowdisplayskip{2pt}
\begin{equation}
    p(\mathbf{R} \mid \pi, \bar{\mathbf{R}}, \mathbf{\Sigma}) = \sum_i \pi_i ~ \mathcal{N}_{SO(3)}(\bar{\mathbf{R}}_i, \mathbf{\Sigma}_i)
\vspace{-0.3em}
\end{equation}
\endgroup
where $\pi_i \in\mathbb{R}$ is the mixture coefficient for the i-th $\mathcal{N}_{SO(3)}$ component.

%As quaternions $\mathbf{q}$ and rotation matrices $\mathbf{R}$ are interchangeable representations the Concentrated Gaussian distribution in SO(3) is implemented on a quaternion basis.
\rebuttal{We want to emphasize that the presented concept on differentiable probabilistic (residual) rotations is applicable to a wide range of problems exceeding motion prediction.}

{\bf Typed Graph Attention.}
To make use of the information of the human skeleton, the entire model is comprised of two building blocks which preserve and efficiently resemble the skeleton structure. Both modules utilize a Graph Influence Matrix $G \in \mathbb{R}^{N \times N}$ inspired by previous work \cite{Kipf2017Semi-SupervisedNetworks, Li2020DynamicPrediction, Mao2019LearningPrediction}. Matrix multiplying the Graph Influence Matrix with a Graph State Matrix $\textbf{x} \in \mathbb{R}^{N \times D_I}$ calculates an element-wise weighted sum of each node's state. This operation is known as \textit{Graph Convolution} \cite{Kipf2017Semi-SupervisedNetworks} or \textit{Graph Attention} \cite{Velickovic2017GraphNetworks} where the attention weights are learned instead of inferred from node states.

To allow for model particularities depending on the semantic class $S_i$ of node $n_i$, we define a typed weight tensor $^{N}\mathbf{W} \in \mathbb{R}^{N\times D_I \times D_O}$ as $N$ stacked weight matrices $\textbf{W}_S \in \mathbb{R}^{D_I \times D_O}$ where $\textbf{W}_S$:
\begingroup
\setlength\abovedisplayskip{4pt}
\setlength\belowdisplayskip{4pt}
\begin{equation}
^{N}\mathbf{W} = 
\begin{bmatrix}
\textbf{W}_{S_0} &
\textbf{W}_{S_1} &
\hdots &
\textbf{W}_{S_N}
\end{bmatrix}
\end{equation}
\endgroup
We define the multiplication operator $\cdot$ as a batched matrix multiplication between the typed weight tensor $^{N}\mathbf{W}$ and the graph input matrix $\mathbf{x} \in \mathbb{R}^{N\times D_I}$
\begingroup
\setlength\abovedisplayskip{4pt}
\setlength\belowdisplayskip{4pt}
\begin{equation}
f(\mathbf{x}) = {}^{N}\mathbf{W} \cdot \mathbf{x}
\end{equation}
\endgroup
All nodes $n_i$ of the same type $S_i$ share the same weights and all $N$ nodes are processed with a single batched matrix multiplication allowing for efficient learning.

{Typed Graph (TG)-Linear:} Using both concepts, attention and typed weights, we define the equivalent to a linear fully connected layer in our graph neural network as
\begingroup
\setlength\abovedisplayskip{4pt}
\setlength\belowdisplayskip{4pt}
\begin{equation}
f(\mathbf{x}) = \mathbf{G}  (^{N}\mathbf{W} \cdot x)
\end{equation}
\endgroup

{Typed Graph (TG)-GRU:}
To capture temporal dependencies within the model we introduce the typed graph equivalent to an GRU layer as
\begingroup
\setlength\abovedisplayskip{3pt}
\setlength\belowdisplayskip{4pt}
\begin{align*}
\begin{split}
r_t &= \sigma_g(\mathbf{G}_t (^{N}\mathbf{W}_{r} \cdot x_t) + \mathbf{G}_t (^{N}\mathbf{U}_{r} \cdot h_{t-1}) + b_f) \\
z_t &= \sigma_g(\mathbf{G}_t (^{N}\mathbf{W}_{z} \cdot x_t) + \mathbf{G}_t (^{N}\mathbf{U}_{z} \cdot h_{t-1}) + b_f) \\
n_t &= \sigma_g(\mathbf{G}_t (^{N}\mathbf{W}_{n} \cdot x_t) + r_t \circ \mathbf{G}_t (^{N}\mathbf{U}_{n} \cdot h_{t-1}) + b_f) \\
h_t &= (1 - z_t) \circ n_t + z_t \circ h_{t-1}\\
\mathbf{G}_t &= \mathbf{G}_{t-1} + \mathbf{G}_{ta}
\end{split}
\end{align*}
\endgroup

where $\mathbf{G}_{0}$,  $^{N}\mathbf{W}$ and $^{N}\mathbf{U}$ are trainable parameters. $h$ is the GRUs state and $\sigma$ represents an activation function. The input $x \in \mathbb{R}^{N\times D_I}$ holds $D_I$ dimensional information on the $N$ nodes. The Graph Influence Matrix $\mathbf{G}$ is initialized as unit matrix and is optimized during training. For the \textit{TG-GRU} an additional Temporal Additive Graph Influence Matrix $G_{ta} \in \mathbb{R}^{N \times N}$ is initialized as a zero matrix and is optimized to capture the temporal change of influence between nodes over time.

\section{\algname{}} \label{sec:model}

\begin{figure}
    \centering
    \includegraphics[width=\linewidth]{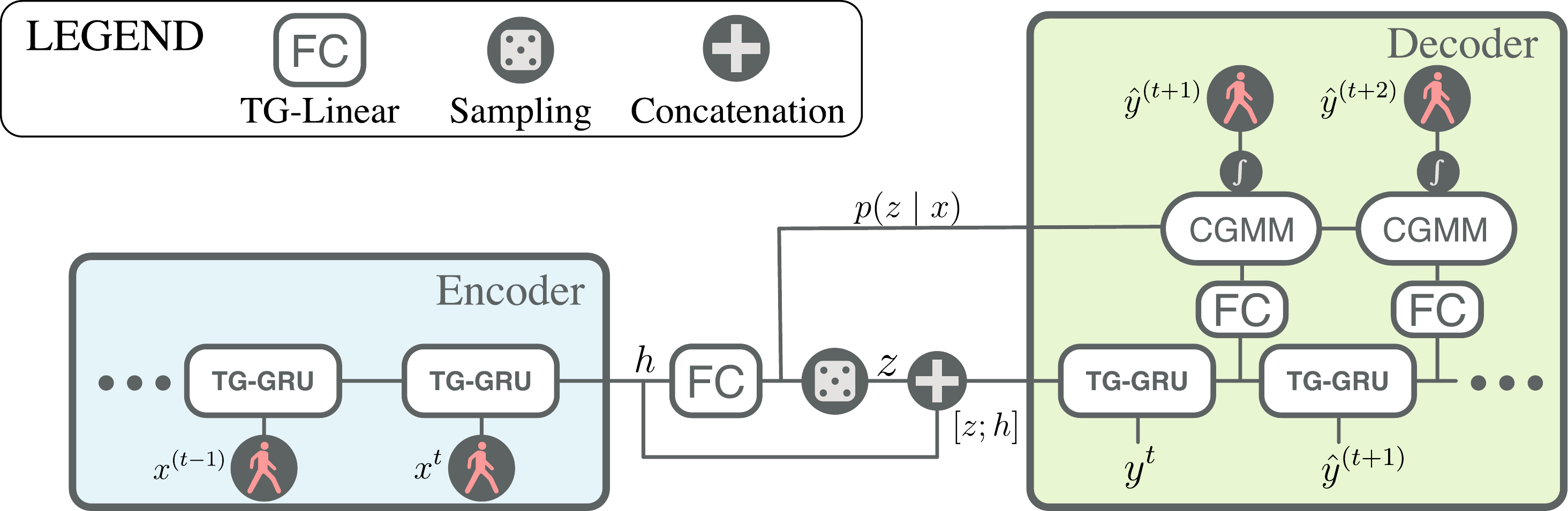}
    \caption{Our network architecture: The encoder abstracts human's historic poses into a hidden representation $h$ using a \textit{TG-GRU}. This representation is used to infer the distribution over the latent variable $p(z \mid \mathbf{x})$, and is fed into the decoder together with latent samples $z$. The decoder, again, uses a \textit{TG-GRU} to compute the output distribution. Notably, $p(z \mid \mathbf{x})$ is reused as mixing coefficients in the output distribution.}
    \label{fig:architecture}
\end{figure}

Our model\footnote{All of our source code, trained models, and data can be found online at \url{https://github.com/TUM-AAS/motron-cvpr22}.} 
is visualized in \cref{fig:architecture}. From a high-level perspective, we combine latent discrete variables with probabilistic mixture distribution as output structure to model the diverse nature of human motion while embedding the skeleton graph structure directly into the learning and inference procedure by using only \textit{Typed Graph} components. 

This model extends core concepts of probabilistic, multimodal deep learning towards the application of human motion prediction. We call our model \emphalgname.

{\bf Graph Structure Embodiment.} We enable efficient use of graph-structured information by imbuing the architecture from end to end. Thus, the architecture is fully described by the two building blocks \textit{TG-Linear} and \textit{TG-GRU}; there is no fully connected influence between hidden node states. This leads to natural modeling of the information flow, where weights are shared between symmetric joints as well as a reduction in model parameters of about 40\%.

{\bf Modeling Motion History.} 
Starting from the input representation $\mathbf{x} = \mathbf{s}_{1,\dots,N}^{(t - H : t)} \in \mathbb{R}^{H \times N \times D}$, the model needs to encode a node's current state and its history. To encode the observed history of the node, its current and previous states are fed into a \textit{TG-GRU} network. The final output is encoded to the model's hidden state $h$ by a \textit{TG-Linear} layer. 
%For the entire \textit{Encoder} we employ weight sharing between all node types as well as limiting $\mathbf{G}$ to the identity matrix. We found this to stabilize training and improve performance as the same hidden representation is enforced for all nodes, making it straightforward for the \textit{Decoder} to combine node states element-wise.

{\bf Explicitly Accounting for Multimodality.} \emphalgname{} explicitly handles multimodality by leveraging a probabilistic latent variable architecture. % loosely inspired by \textit{(Conditional) Variational Autoencoder}s ((C)VAEs). 
It produces the target distribution $p(\mathbf{y} \mid \mathbf{x})$ by introducing a discrete latent variable $z$, 
\begingroup
\setlength\abovedisplayskip{4pt}
\setlength\belowdisplayskip{4pt}
\begin{equation}
    p(z \mid x) = \frac{1}{N} \sum_i f_{TG-Linear}(h)_i,
\vspace{-0.3em}
\end{equation}
\endgroup
which encodes high-level latent behavior and allows for $p(\mathbf{y} \mid \mathbf{x})$ to be expressed as
\begingroup
\setlength\abovedisplayskip{4pt}
\setlength\belowdisplayskip{4pt}
\begin{equation}\label{eqn:model}
    p(\mathbf{y} \mid \mathbf{x}) = \sum_z p_\psi(\mathbf{y} \mid \mathbf{x}, z) p_\theta(z \mid \mathbf{x}),
\vspace{-0.3em}
\end{equation}
\endgroup
where $\psi$ and $\theta$ are deep neural network weights that parameterize their respective distributions.
\rebuttal{Unlike latent variables in probabilistic variational (e.g. (C)VAEs), this latent variable is not explicitly encouraged to learn a representation for the decoder (e.g. using KL-divergence), but implicitly enables the decoder to differentiate between modes.}

{\bf Producing Motions.}\label{sec:prod_motions} The sampled latent variable $z$ is concatenated with the hidden representation vector $h$ and fed into our \textit{TG-GRU} decoder. Each \textit{TG-GRU} cell outputs the parameters $\bar{\mathbf{R}}, \mathbf{\Sigma}$ of a $\mathcal{N}_{SO(3)}$ for each node. Using $\pi_i = p(z = i \mid \mathbf{x})$ we produce the mixture model $\dot{\mathbf{q}} \sim \mathcal{N}^{\pi}_{SO(3)}$ over differential quaternions as an intermediate output distribution.

Thus, $z$ being discrete is necessary as it enables us to rethink \cref{eqn:model} as a $\mathcal{N}^{\pi}_{SO(3)}$ with mixture distribution $p(z \mid x)$. It also aids in interpretability, as one can visualize the high-level behaviors resulting from each $z$ by sampling motions (see \cref{fig:qualitative}).

Using the closed composition formula \cref{eq:comp} of the Concentrated Gaussian in SO(3) we can ''integrate'' the distribution over differential quaternions to the final output distribution over quaternions. The intermediate step is necessary as motion samples are produced by sampling differential quaternions and subsequently ''integrating'' them to motions. Directly sampling the output distribution would lead to time inconsistent motions. Additionally, using differential output for recurrent layers is known to ease the learning problem and improve convergence \cite{Salzmann2020Trajectron, Ivanovic2018Generative}.

{\bf Training the Model.} 
Commonly, learning good representations for probabilistic latent variables is achieved by including the ground truth $y$ as input to the latent layer during training and simultaneously introducing a \textit{Kullback–Leibler Divergence} (KL) loss term to \textit{squeeze} out the dependency on $y$ during the training process \cite{Higgins2016VAE:Framework, Salzmann2020Trajectron}. When using the CVAE framework, these competing conditions can lead to unstable training behavior and the collapse of the latent distribution (KL divergence towards zero). 
In contrast, we do not input the ground truth but give the model the option to use the latent capacity $p_\theta(z \mid \mathbf{x})$ to maximize \cref{eqn:model} in \cref{eqn:loss_fn}.
%To further encourage this behaviour we adapt the mutual information maximization loss term \cite{Zhao2019InfoVAE:Autoencoders}, and modify it to use discrete latent states.
Formally, we aim to solve
\begingroup
\setlength\abovedisplayskip{4pt}
\setlength\belowdisplayskip{4pt}
\begin{equation}\label{eqn:loss_fn}
\begin{aligned}
\max_{\theta, \psi} \sum_{i=1}^N\ & \mathbb{E}_{z \sim p_\theta(\cdot \mid \mathbf{x}_i)} \big[\log p_\psi(\mathbf{y}_i \mid \mathbf{x}_i, z)\big]\\ 
\end{aligned}
\vspace{-0.3em}
\end{equation}
\endgroup
Notably, no reparameterization trick, commonly needed for training probabilistic latent variable models \cite{Kingma2013Auto-EncodingBayes, Jang2017CategoricalGumbel-Softmax} is used to backpropagate through the categorical latent variable $z$ as it is not sampled during training time. Instead, \cref{eqn:loss_fn} is directly computed since the latent space has only $|Z|$ discrete elements. (For an in depth discussion see \cref{sec:GS_rep})

{\bf Output Configurations.} Based on the desired use case, \emphalgname{} can produce many different outputs. The main four are outlined below.
\begin{enumerate}
    \item \emph{Distribution}: Due to the use of a discrete latent variable and probabilistic output structure, the model can provide an analytic output distribution by directly computing~\cref{eqn:model}. This parametric $\mathcal{N}^{\pi}_{SO(3)}$ distribution entails the complete information inferred by the model.
    
    \item \emph{Sampled}: The model's sampled output, where $z$ and $y$ are sampled sequentially according to 
    \begingroup
\setlength\abovedisplayskip{3pt}
\setlength\belowdisplayskip{0pt}
    \begin{equation}
    \begin{aligned}
    z \sim p_\theta(z \mid \mathbf{x}), \hspace{0.5cm} \mathbf{y} \sim p_\psi(\mathbf{y} \mid \mathbf{x}, z).
    \end{aligned}
    \end{equation}
    \endgroup
    
    %\item \emph{Most Likely (ML)}: 
    %The model's deterministic and most-likely single output.
    %The motion with the highest log-likelihood under the full output distribution
    %\begingroup
%\setlength\abovedisplayskip{3pt}
%\setlength\belowdisplayskip{0pt}
%    \begin{equation}
%    \begin{aligned}
%    \mathbf{y} &= \arg \max_{\mathbf{y}} \sum_z p_\theta(z \mid \mathbf{x}) p_\psi(\mathbf{y} \mid \mathbf{x}, z).
%    \end{aligned}
%    \end{equation}
%    \endgroup
    
    \item \vspace{-0.5em} \emph{Most Likely Mode (ML-Mode)}: 
    The model's deterministic and most-likely single output. % both in position and configuration space. 
    %Mean deterministic motions from the other latent modes are possible too.
    The high-level latent behavior mode and output trajectory are the modes of their respective distributions, where
    \begingroup
\setlength\abovedisplayskip{3pt}
\setlength\belowdisplayskip{0pt}
    \begin{equation}
    \begin{aligned}
    z_\text{mode} &= \arg \max_z p_\theta(z \mid \mathbf{x}), \\
    \mathbf{y} &= \arg \max_{\mathbf{y}} p_\psi(\mathbf{y} \mid \mathbf{x}, z_\text{mode}).
    \end{aligned}
    \end{equation}
    \endgroup
    
    \item \vspace{-0.5em} \emph{Weighted Mean (W-Mean)} 
    The mean of all latent modes weighted by their probability. Following \cite{Markley2007AveragingQuaternions}, it is given as the normalized largest eigenvector of $QQ^T$ where $Q$ is a matrix of stacked quaternion column vectors, where
    \begingroup
    \setlength\abovedisplayskip{3pt}
    \setlength\belowdisplayskip{0pt}
    \begin{equation}
    \begin{aligned}
        \mathbf{q}_n = \mathbf{y}_n &= \arg \max_{\mathbf{y}} p(\mathbf{y} \mid \mathbf{x}, z=n), \\
        Q &= [\pi_1 \mathbf{q}_{1}, ..., \pi_{|Z|} \mathbf{q}_{|Z|} ].
    \end{aligned}
    \end{equation}
    \endgroup
    
    %\item \emph{Deterministic} Restricting the number of latent states to one $|Z| = 1$, and replacing \cref{eqn:loss_fn} with the mean absolute loss function we can switch our model towards producing a single deterministic motion.
    
\end{enumerate}

\vspace{-1em}
\section{Experiments}
While desiderata (\rom{3}) and (\rom{5}) in \cref{sec:Intro} are explicitly fulfilled by the embedded human structure and the use of $\mathcal{N}^{\pi}_{SO(3)}$ as output distribution, we conduct both quantitative and qualitative experiments to show that our method also succeeds in the remaining desiderata. 

We, therefore, structure our experiments as follows:
First, we show that \emphalgname{} performs best in capturing humans' probabilistic and diverse nature (\rom{2}) by introducing a new probabilistic metric to the field of human motion prediction.
Secondly, we highlight the learned multimodal structure (\rom{1}) by evaluating deterministic outputs of high likelihood modes. Later we also give a visualization of how these modes manifest in distinct motions.
Finally, we show that our approach can handle incomplete data (\rom{4}) in the form of occluded joints.

{\bf Datasets.} We provide quantitative experimentation results on two datasets; namely the \textit{Human 3.6 Million} (H3.6M) \cite{Ionescu2014Human3.6M:Environments} dataset and the \textit{Archive of Motion Capture as Surface Shapes} (AMASS) \cite{Mahmood2019AMASS:Shapes}. AMASS is a unified collection of 18 motion capture datasets totaling 13944 motion-sequences from 460 subjects performing a large variety of actions. In comparison, the H3.6M dataset consists of 240 sequences from 8 subjects performing 15 actions each. 

%As deterministic and generative algorithm historically differ in their evaluation methodology we split this section to evaluate the two qualities independently before drawing a joint conclusion.

\subsection{\rebuttalrm{Generative} \rebuttal{Probabilistic} Evaluation}
{\bf Metrics.}
\rebuttalrm{Generative} \rebuttal{Probabilistic} approaches have been compared on the basis of a variety of metrics in position space, most prominent are Best-of-N metrics where $N=50$ motions are sampled from the\rebuttalrm{generative} model; the best metric value of these is reported. Such metrics, however, only present limited insights on a model's output distribution as only a single motion of an arbitrary number $N$ is evaluated. 
To fully assess an algorithm's probabilistic capabilities, we propose an alternative evaluation methodology for\rebuttalrm{generative} \rebuttal{probabilistic} algorithms where we measure the ability to accurately capture and reproduce the underlying uncertainty distribution of motions. To this point we adopt the \textit{KDE-NLL} metric \cite{Ivanovic2019TheGraphs} to assess the method's Negative Log-Likelihood (NLL) by fitting a probability distribution, using Kernel Density Estimate (KDE) \cite{Parzen1962OnMode}, to output samples. Although \algname{} can compute its own log-likelihood, we apply the same evaluation methodology to maintain a directly comparable performance measure. As the NLL is unbounded, we clip it to a maximum value of $20$ ($\sim 2*10^{-7}\%$) in order to prevent single outliers from dominating.
Still, to be comparable, we additionally follow the evaluation methodology of Yuan \etal \cite{Yuan2020DLow:Prediction}: We report the \textit{Average Pairwise Distance} (APD) as a measure of sample diversity as well as the Best-of-N metrics \textit{Average Displacement Error} (ADE), and \textit{Final Displacement Error} (FDE) as measures of quality. Further, we report their Multi-Modal ADE and FDE metrics (MMADE and MMFDE). ``Similar`` motions are grouped by using an arbitrary distance threshold at $t=0$ and the average metric over all these grouped motions is reported. As close poses at a single instance can, however, belong to entirely different motions (\cref{sec:mm_metric}), the KDE-NLL reports a more holistic representation of a model's\rebuttalrm{generative} \rebuttal{probabilistic} capabilities.

{\bf\rebuttalrm{Generative} \rebuttal{Probabilistic}  Baselines.}
We focus on the current state-of-the-art\rebuttalrm{generative} algorithm (1)~\textit{DLow} \cite{Yuan2020DLow:Prediction} to compare our desiderata side to side. \textit{DLow} uses an adapted VAE algorithm to generate samples without collapsing to a single mode.
For the standard\rebuttalrm{generative} \rebuttal{probabilistic} experiment methodology, we further report methods based on CVAEs: (2)~\textit{Pose-Knows} \cite{Walker2017TheFutures} and (3)~\textit{MT-VAE} \cite{Yan2018MT-VAE:Dynamics};  GAN based (4)~\textit{DeLiGAN} \cite{Gurumurthy2017DeLiGAN:Data} and diversity promoting methods (5)~\textit{Best-of-Many} \cite{Bhattacharyya2018AccurateObjective}, (6)~\textit{GMVAE} \cite{Dilokthanakul2017DeepAutoencoders}, and (7)~\textit{DSF} \cite{Yuan2020DiverseProcesses}. We were not able to compare to \textit{LCP-VAE} \cite{Aliakbarian2021ContextuallyPrediction} for a lack of open accessible source code.

{\bf Evaluation Methodology.}
In order to compare to other\rebuttalrm{generative} \rebuttal{probabilistic} methods, which output motions solely in position space, we use the forward kinematic of the respective test subject to convert our joint configuration samples to joint positions. Predicting in configuration space and using human's forward kinematic to produce motions in position space, ensures our motions to be kinematic feasible. However, they bring a disadvantage compared to approaches directly outputting joints positions, common for\rebuttalrm{generative} \rebuttal{probabilistic} approaches, as they are not constrained by a rigid bone structure (\cref{sec:bone_deform}).
For the KDE-NLL metric, we sample $N=1000$ motions from each method and fit a KDE for each future timestep.
For the $N=50$ metrics (APD, ADE, FDE, MMADE, MMFDE) we use 50 motions, sampled from our intermediate output distribution over $\dot{q}$ and ''integrated''; these motions are transformed into position space using the respective subjects forward kinematics.
We train the model on 0.5 seconds of history and predict a two second horizon. Generally, we can dynamically change the prediction horizon online thanks to the flexible decoder structure during inference.

\begin{figure}
    \centering
    \includegraphics[trim=0.3cm 0.65cm 0.3cm 0.7cm, width=\linewidth]{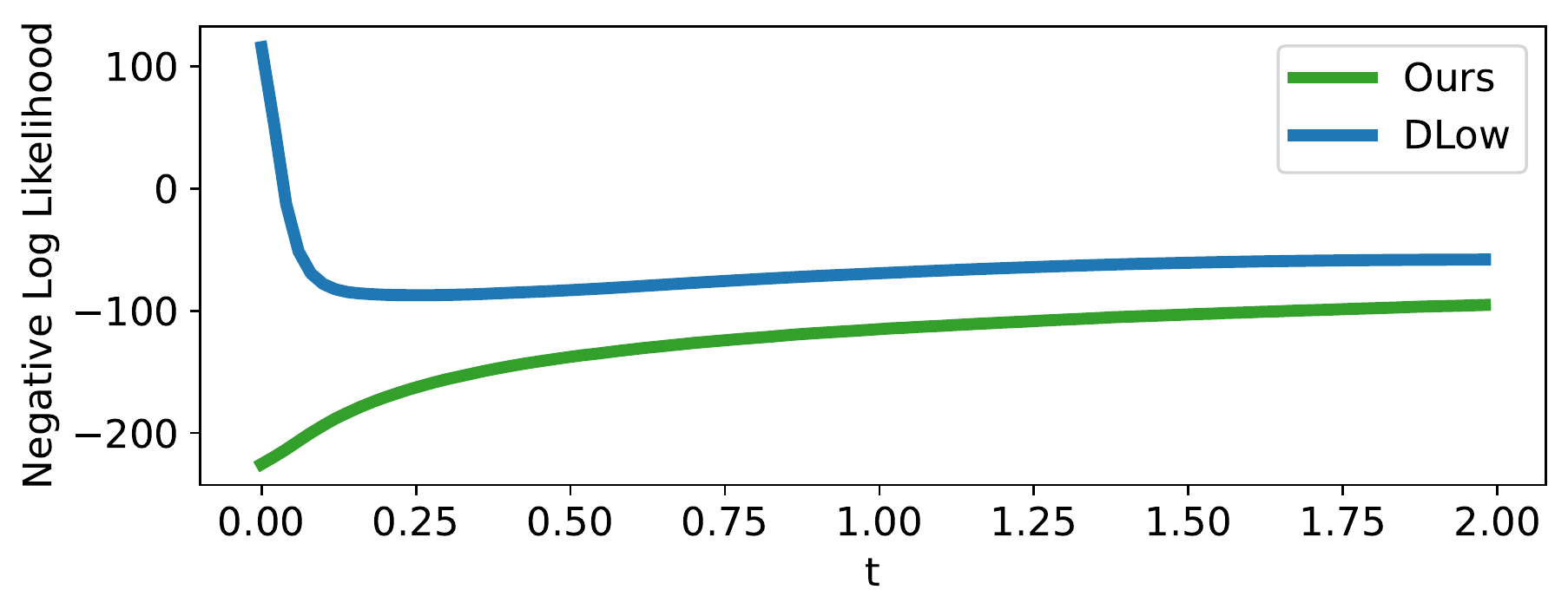}
    \caption{Negative Log Likelihood of the ground truth in the fitted KDE distribution of samples from each model. Lower is better. Samples from \textit{DLow} are over-confidently wrong for early prediction timesteps.}
    \label{fig:nll}
\end{figure}

{\bf Results.}
\cref{fig:nll} shows that \emphalgname{} clearly outperforms the current state-of-the-art algorithm \textit{DLow} in representing the underlying motion distribution of the human subject and therefore having a lower NLL over all prediction timesteps. This result is supported by the Best-of-N metrics over one and two seconds in \cref{tab:gen}. We show significantly better results than \textit{DLow} on ADE and FDE. Notably, this is achieved while using less diverse motion samples (lower APD) indicating that our samples are concentrated around likely motions. MMADE and MMFDE values for different thresholds are presented in \cref{fig:mm_metric_plot} in \cref{sec:mm_metric}. For smaller thresholds we outperform \textit{DLow} while for higher thresholds, where possibly uncorrelated motions are evaluated together as displayed in \cref{fig:mm_metric_ex} in \cref{sec:mm_metric}, we unsurprisingly achieve lower scores as our approach does not over-diversify its predictions.

\begin{table}
\fontsize{8}{8}\selectfont
\renewcommand\arraystretch{1.2}
\renewcommand\tabcolsep{1pt}
\begin{tabular*}{\linewidth}{@{\extracolsep{\fill}}lccc|ccc@{}}
 & \multicolumn{3}{c}{1 s} & \multicolumn{3}{c}{2 s} \\ \hline
 & APD & ADE & FDE & APD & ADE & FDE \\ \hline
DSF \cite{Yuan2020DiverseProcesses} & - & - & - & 9.330 & 0.493 & 0.592 \\
DeLiGAN \cite{Gurumurthy2017DeLiGAN:Data} & - & - & - & 6.509 & 0.483 & 0.534 \\
GMVAE \cite{Dilokthanakul2017DeepAutoencoders} & - & - & - & 6.769 & 0.461 & 0.555 \\
Best-of-Many \cite{Bhattacharyya2018AccurateObjective} & - & - & - & 6.265 & 0.448 & 0.533 \\
%HP-GAN & 7.214 & 0.858 & 0.867 & 0.847 & 0.858 \\
MT-VAE \cite{Yan2018MT-VAE:Dynamics} & - & - & - & 0.403 & 0.457 & 0.595 \\
Pose-Knows \cite{Walker2017TheFutures} & - & - & - & 6.723 & 0.461 & 0.560 \\
%acLSTM & 0 & 0.789 & 1.126 & 0.849 & 1.139 \\
%ERD & 0 & 0.722 & 0.969 & 0.776 & 0.995 \\
DLow \cite{Yuan2020DLow:Prediction} & 5.180 & 0.305 & 0.419 & 11.741 & 0.425 & 0.518 \\
\hline
Ours & 3.453 & \textbf{0.252} & \textbf{0.350} &  7.168 & \textbf{0.375} & \textbf{0.488}
\end{tabular*}
\caption{Best of $N=50$ evaluation against\rebuttalrm{generative} \rebuttal{probabilistic} algorithms on a prediction horizon of one and two seconds on H3.6M dataset.}
\label{tab:gen}
\end{table}

\vspace{-0.4em}
\subsection{Deterministic Evaluation}\label{sec:det_exp}

{\bf Metrics.}
We report the \textit{Mean Angle Error} (MAE-L2) as the Euclidean distance of the stacked (ZYX-)Euler angles as well as the \textit{Mean Per Joint Position Error} (MPJPE) \cite{Ionescu2014Human3.6M:Environments} which is calculated using the human's skeleton forward kinematic. Those are the standard evaluation metrics in deterministic motion prediction \cite{Mao2020HistoryAttention, Mao2019LearningPrediction, Li2020DynamicPrediction, Jain2016Structural-RNN:Graphs, Pavllo2019ModelingNetworks}. To better understand the influence of the learned latent multimodality we apply an \textit{Best-of-N} evaluation for the MAE-L2 and MPJPE metric. Here we report the value for the motion with the best average metric value over all prediction timesteps originating from the $N$ modes' means with the highest probability $p(z \mid x)$.

%We compare our method to an exhaustive set of state-of-the-art deterministic and generative approaches.

{\bf Deterministic Baselines.} We evaluate against the following deterministic approaches: (1)~\textit{Zero Velocity}: All joints keep their current state at prediction time throughout the entire prediction horizon. (2)~\textit{GRU sup.} \cite{Martinez2017OnNetworks}: Simple encoder-decoder structure using GRUs for variable history and prediction horizon and exponential maps as data representation. (3)~\textit{Quaternet} \cite{Pavllo2019ModelingNetworks}: Encoder-Decoder architecture using GRUs and quaternion data representation. (4)~\textit{HistRepItself} \cite{Mao2020HistoryAttention}: Transformer \cite{Vaswani2017AttentionNeed} encoder and fixed prediction horizon graph convolution decoder. Data is pre-processed using the \textit{Discrete Cosine Transform} (DCT) on the time dimension and the model predicts a residual before the output is transformed back using the inverse DCT. \rebuttal{(5)~\textit{ST-Transformer} \cite{Aksan2021APrediction}: Decoupled temporal and spatial self-attention.}
For the H3.6M dataset we state the reported values of (2) - (3) and re-run the evaluation of (4) as they originally did not account for $2\pi$ angle discontinuity.
For the AMASS dataset we re-train (4) \textit{HistRepItself} to match our test split.
%Some approaches \cite{Mao2020HistoryAttention} report results for configuration space metrics (MAE-L2) and positions space metrics (MPJPE) from different models, where the MPJPE model is directly trained on 3-D joint positions. While this shows improvement on MPJPE it also leads to physically unfeasible motions as the rigid bone structure is deformed (\cref{sec:bone_deform}); leading to an unfair advantage. Thus, we use a forward kinematic approach for all compared deterministic models to calculate the MPJPE metric.
We were not able to compare to some other methods for a lack of open source code (\textit{AGED} \cite{Gui2018AdversarialPrediction}) or their computational complexity (\textit{DMGNN} \cite{Li2020DynamicPrediction} - 62 Million parameter)

{\bf Evaluation Methodology.}
It has become common to benchmark on 8 fixed sequences per action of a single test subject on the H3.6M dataset. This has been shown to be un-representative \cite{Pavllo2019ModelingNetworks}. Thus, we report results on 8 \cite{Mao2020HistoryAttention, Li2020DynamicPrediction, Pavllo2019ModelingNetworks, Martinez2017OnNetworks} (see \cref{sec:add_det_res}) and 256 \cite{Pavllo2019ModelingNetworks, Mao2020HistoryAttention} samples per action on the H3.6M data to be comparable to past methods.
For the AMASS dataset, we report results on the official test split, consisting of the \textit{Transitions} and \textit{SSM} dataset. While prior authors \cite{Mao2020HistoryAttention} have argued that the \textit{Transitions} dataset is not suitable for evaluating prediction algorithms for their change of action within sequences, we argue that such behavior can happen in real-world applications.
We subsample the H3.6M dataset to 25 HZ and the AMASS dataset to 20 HZ as most of the included datasets have been recorded with a framerate divisible by 20 but not by 25. \rebuttal{When comparing to \textit{ST-Transformer} \cite{Aksan2021APrediction} (\cref{tab:amass_st}) we follow their evaluation methodology.}
For the H3.6M dataset, we report metric values previously published on both 8 and 256 samples per action. As the AMASS dataset has been released recently, there are not many published results, yet. Thus, we retrain the current best state-of-the-art algorithm \textit{HistRepItself} and re-train it on the official test split.
We train the model on two seconds of history and predict one second into the future.

{\bf Results.}
We evaluate our approach on common single sample metrics against fully deterministic approaches. \cref{tab:det_h36m} summarizes the results on the H3.6M dataset. Even though we don't explicitly optimize for a deterministic output, our Weighted-Mean output outperforms all other state-of-the-art algorithms. 

We want to point out that we introduced the W-Mean output configuration solely for these metrics commonly used in deterministic evaluation. This allows us to point to the shortcomings of both, deterministic algorithms and their corresponding evaluation metrics: They produce motions which represent the average of all likely motions given the motion history. This average motion, however, may represent a unlikely or even infeasible motion. This (unintentional) behavior is followed by our W-Mean output configuration. Thus, we want to emphasize that while we outperform other algorithms on specific metrics we advise against using the W-Mean output configuration for actual applications. 

The lower rows of \cref{tab:det_h36m} and \cref{tab:det_amass} supports that our approach captures multimodality by committing to a specific motion per mode; thereby attributing appropriate probability mass to less likely motions. As such, the most likely deterministic mode (ML-Mode) commits to the mode which is best explained by the data. However, on average it accumulates a higher error compared to W-Mean as it is expected to represent a ``wrong`` motion with probability $p=1 - \text{max}_i(\pi_i)$ (see \cref{sec:prod_motions}). Looking at the set of likely deterministic mode outputs (BoN in \cref{tab:det_h36m} and \cref{tab:det_amass}), it becomes clear that one of the motion modes performs exceptionally better than the mean output of deterministic regressors. This behavior is further visualized for a single example in \cref{sec:qualitative}.

\begin{table}
\fontsize{8}{8}\selectfont
\renewcommand\arraystretch{1.2}
\renewcommand\tabcolsep{2pt}
\begin{tabular*}{\linewidth}{@{\extracolsep{\fill}}lcccccccc}
 & \multicolumn{8}{c}{MAE (L2)} \\
milliseconds & 80 & 160 & 320 & 400 & 560 & 720 & 880 & 1000  \\ \hline
Zero Vel. & 0.40 & 0.70 & 1.11 & 1.25 & 1.46 & 1.63 & 1.76 & 1.84 \\
GRU sup. \cite{Martinez2017OnNetworks} & 0.43 & 0.74 & 1.15 & 1.30 & - & - & - & - \\
Quarternet \cite{Pavllo2019ModelingNetworks} & 0.37 & 0.62 & 1.00 & 1.14 & - & - & - \\
HistRepItself \cite{Mao2020HistoryAttention} & \textbf{0.28} & 0.52 & 0.88 & 1.02 & 1.23 & \textbf{1.40} & 1.55 & 1.64 \\
\hline
%Ours Det. & 0.30 & 0.54 & 0.89 & 1.03 & \textbf{1.23} & \textbf{1.39} & \textbf{1.53} & 1.61 \\
Ours W-Mean & \textbf{0.28} & \textbf{0.51} & \textbf{0.87} & \textbf{1.01} & \textbf{1.22} & \textbf{1.40} & \textbf{1.54} & \textbf{1.63} \\
\hline
Ours ML-Mode & 0.28 & 0.51 & 0.88 & 1.02 & 1.24 & 1.42 & 1.58 & 1.67 \\
Ours Bo3-Modes & 0.28 & 0.50 & 0.85 & 0.97 & 1.16 & 1.31 & 1.45 & 1.54 \\
Ours Bo5-Modes & 0.28 & 0.51 & 0.84 & 0.96 & 1.13 & 1.28 & 1.42 & 1.51 \\
\end{tabular*}
\caption{Average angle error on 256 samples per action on the H3.6M test dataset. A break down by actions and the results on the MPJPE metric can be found in \cref{sec:add_det_res}.}
\label{tab:det_h36m}
\end{table}

\begin{table}
\fontsize{8}{8}\selectfont
\renewcommand\arraystretch{1.2}
\renewcommand\tabcolsep{4pt}
\begin{tabular*}{\linewidth}{@{\extracolsep{\fill}}lcccccc}
 & \multicolumn{6}{c}{MAE (L2)} \\
milliseconds & 100 & 200  & 400 & 600 & 800 & 1000 \\ \hline
Zero Vel. & 0.73 & 1.20 & 1.60 & 1.73 & 1.76 & 1.76 \\
HistRepItself \cite{Mao2020HistoryAttention} & 0.45 & 0.78 & 1.06 & \textbf{1.17} & \textbf{1.27} & \textbf{1.33} \\
\hline
Ours W-Mean & \textbf{0.42} & \textbf{0.76} & \textbf{1.05} & 1.19 & \textbf{1.27} & \textbf{1.33} \\
\hline
Ours ML-Mode & 0.42 & 0.76 & 1.08 & 1.22 & 1.31 & 1.38 \\
Ours Bo3-Modes & 0.42 & 0.74 & 1.01 & 1.12 & 1.20 & 1.28 \\
Ours Bo5-Modes & 0.42 & 0.74 & 0.99 & 1.09 & 1.17 & 1.27 
\end{tabular*}
\caption{Average angle error on 10,000 samples from the AMASS test set. The MPJPE metric can be found in \cref{sec:add_det_res}.}
\label{tab:det_amass}
\end{table}

\begin{table}
\fontsize{8}{8}\selectfont
\renewcommand\arraystretch{1.2}
\renewcommand\tabcolsep{2pt}
\begin{tabular*}{\linewidth}{@{\extracolsep{\fill}}lcccc}
 & \multicolumn{4}{c}{MAE (L2)} \\
milliseconds & 100 & 200 & 300 & 400   \\ \hline
ST-Transformer \cite{Aksan2021APrediction} & 0.178 & 0.291 & 0.395 & 0.490 \\
Ours W-Mean  & \textbf{0.147} & \textbf{0.243} & \textbf{0.335} & \textbf{0.420}  \\ \hline
\end{tabular*}
\caption{\rebuttal{Average angle error using the evaluation procedure in \cite{Aksan2021APrediction}.}}
\label{tab:amass_st}
\end{table}

Notably, \emphalgname{} can capture multimodality and reason probabilistically about humans' future motion while being computationally more efficient then a deterministic regressor. With 1.7 Million parameters we need \textbf{half} the computational power than \textit{HistRepItself} (3.4M parameters) and are significantly more efficient compared to the current state-of-the-art algorithm \textit{DLow} with 7.3 Million parameters. Other graph based models, such as \cite{Li2020DynamicPrediction}, even use 62 Million parameters.

\subsection{Ablation Study}\label{sec:ablation}
In this section, we will show the influence of our contributions on the model's performance as well as justify our design choices quantitatively.

{\bf Number of latent modes.} We ablate the number of latent discrete states which manifest as motion modes in the model's output (\cref{tab:abl_modes}). Four and five modes show the best performance. We use five modes for our approach to give the model more expressiveness when necessary. 

\begin{table}
\fontsize{8}{8}\selectfont
\renewcommand\arraystretch{1.2}
\renewcommand\tabcolsep{2pt}
\begin{tabular*}{\linewidth}{@{\extracolsep{\fill}}lccc|cc}
 & \multicolumn{3}{c}{NLL} &  \multicolumn{2}{c}{MAE (L2)} \\
milliseconds & 400 & 1000 & $\sum$ & 400 & 1000 \\ \hline
$|Z| = 1$ & -160.77 & -106.10 & -4032.58 & 1.05 & 1.68 \\
$|Z| = 2$ & -173.21 & -117.20 & -4320.73 & 1.02 & 1.63 \\
$|Z| = 3$ & -176.25 & -121.86 & -4405.46 & 1.02 & 1.65 \\
$|Z| = 4$ & -176.74 & \textbf{-122.03} & -4418.98 & \textbf{1.01} & \textbf{1.63} \\
$|Z| = 5$ & \textbf{-177.01} & -122.02 & \textbf{-4432.40} & \textbf{1.01} & \textbf{1.63} \\
$|Z| = 6$ & -174.50 & -117.70 & -4340.94 & 1.01 & 1.64 \\
\end{tabular*}
\caption{Negative Log Likelihood (NLL) and MAE (L2) performance using different number of latent modes on H3.6M dataset.}
\label{tab:abl_modes}
\end{table}

{\bf Influence of contributions.} To show the influence of our contributions on the model's performance we ablate them individually in \cref{tab:abl_cont}. We first remove the \textit{Typed Graph} weight sharing scheme and subsequently replace it with One-Hot encoded type information which can recover some performance. Next, we replace the Concentrated Gaussian distribution in SO(3) with a standard Multivariate Normal (Gaussian) Mixture Model (GMM) and subsequently with a Bingham distribution. Unlike our $\mathcal{N}_{SO(3)}$, the MVN is not designed to handle rotations and their special characteristics. It, therefore, has worse results in all metrics. The Bingham distribution, in contrast, is designed for rotations but does not support the composition of rotations. Thus, we can not use differential quaternions as intermediate output making the learning task more complex. Also, samples from a Bingham can only be approximated using computationally and memory expensive algorithms (e.g. Metropolis-Hasting). Finally, we enable gradient flow through the latent variable (see \cref{sec:GS_rep} and \cref{sec:model}), which has a minor negative impact.

\begin{table}
\fontsize{8}{8}\selectfont
\renewcommand\arraystretch{1.2}
\renewcommand\tabcolsep{2pt}
\begin{tabular*}{\linewidth}{@{\extracolsep{\fill}}lccc|cc}
 & \multicolumn{3}{c}{NLL} &  \multicolumn{2}{c}{MAE (L2)} \\
milliseconds & 400 & 1000 & $\sum$  & 400 & 1000 \\ \hline
No Typed Graph & -170.28 & -112.35 & -4264.17 & 1.07 & 1.73 \\
One-Hot & -174.78 & -119.33 & -4372.70 & 1.04 & 1.67 \\
\hline
Gaussian Mixture Model & -158.56 & -102.24 & -3879.12 & 1.06 & 1.69 \\
Bingham & -162.58 & -107.52 & -3983.10 & 1.07 & 1.67 \\
\hline
Latent Grad. Flow & -174.68 & -119.99 & -4374.77 & 1.02 & 1.64 \\
\hline
Full & \textbf{-177.01} & \textbf{-122.02} & \textbf{-4432.40} & \textbf{1.01} & \textbf{1.63} \\
\end{tabular*}
\caption{Negative Log Likelihood (NLL) and MAE (L2) performance on ablated model structures on H3.6M dataset.}
\label{tab:abl_cont}
\end{table}

\begin{figure*}
    \centering
    \begin{subfigure}{.185\linewidth}
        \includegraphics[trim=2.1cm 0.8cm 2.1cm 2.7cm, clip, width=\linewidth]{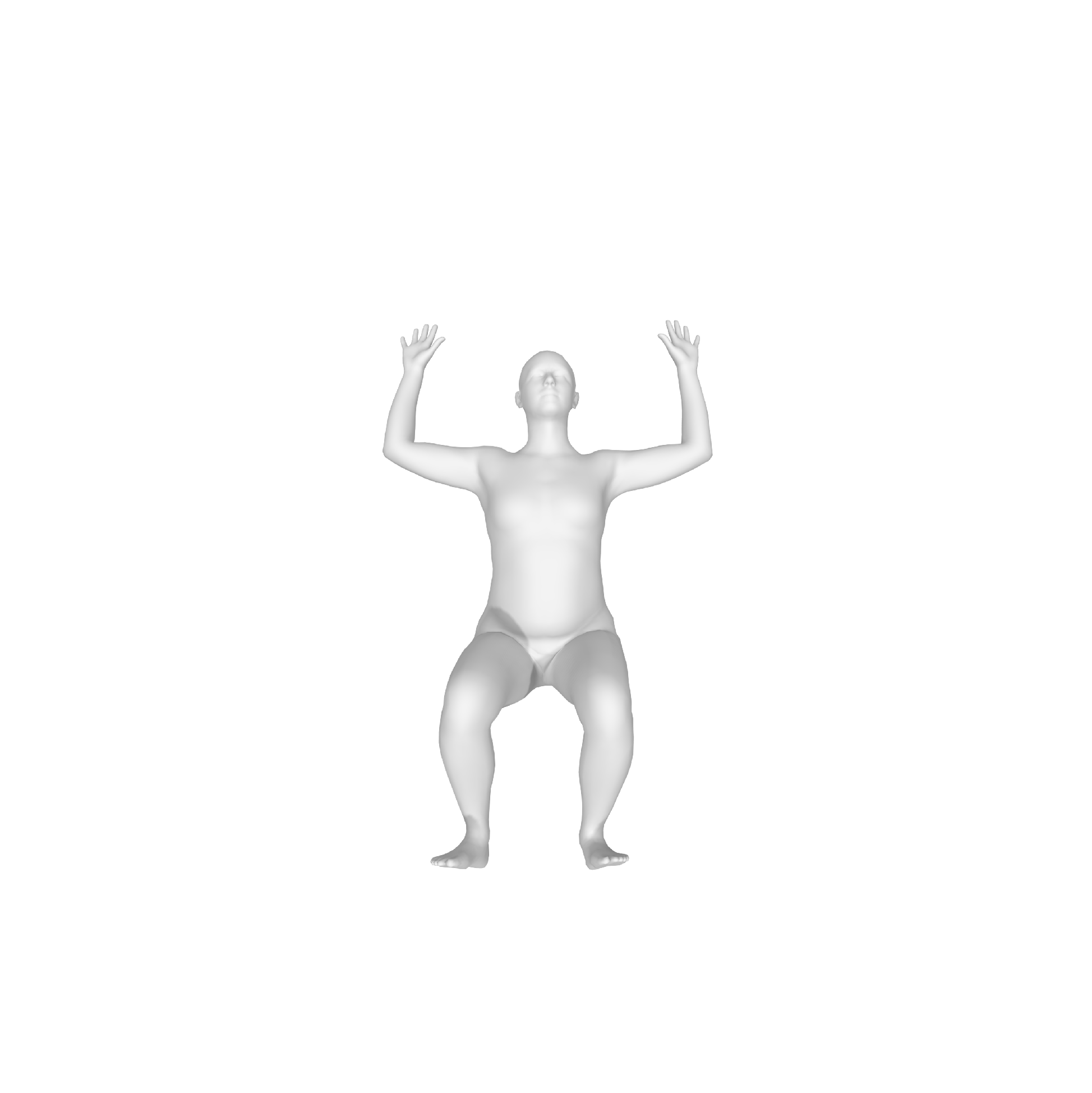}
        \caption{Ground Truth $t=-500ms$}
    \end{subfigure}\hfill
    \begin{subfigure}{.185\linewidth}
        \includegraphics[trim=2.1cm 0.8cm 2.1cm 2.70cm, clip, width=\linewidth]{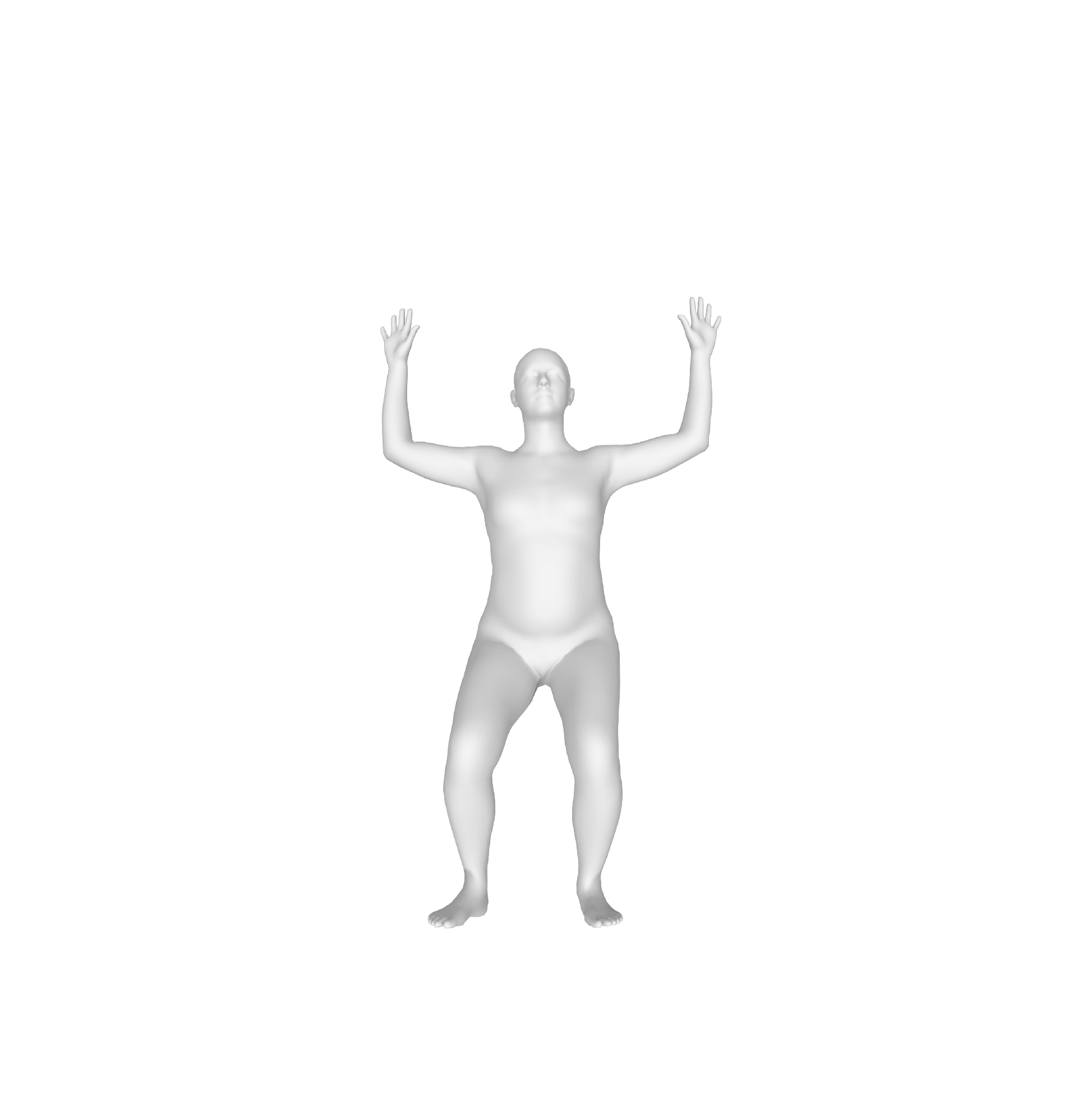}
        \caption{Ground Truth $t=0ms$}
    \end{subfigure}\hfill
    \begin{subfigure}{.185\linewidth}
        \includegraphics[trim=2.1cm 0.8cm 2.1cm 2.7cm, clip,width=\linewidth]{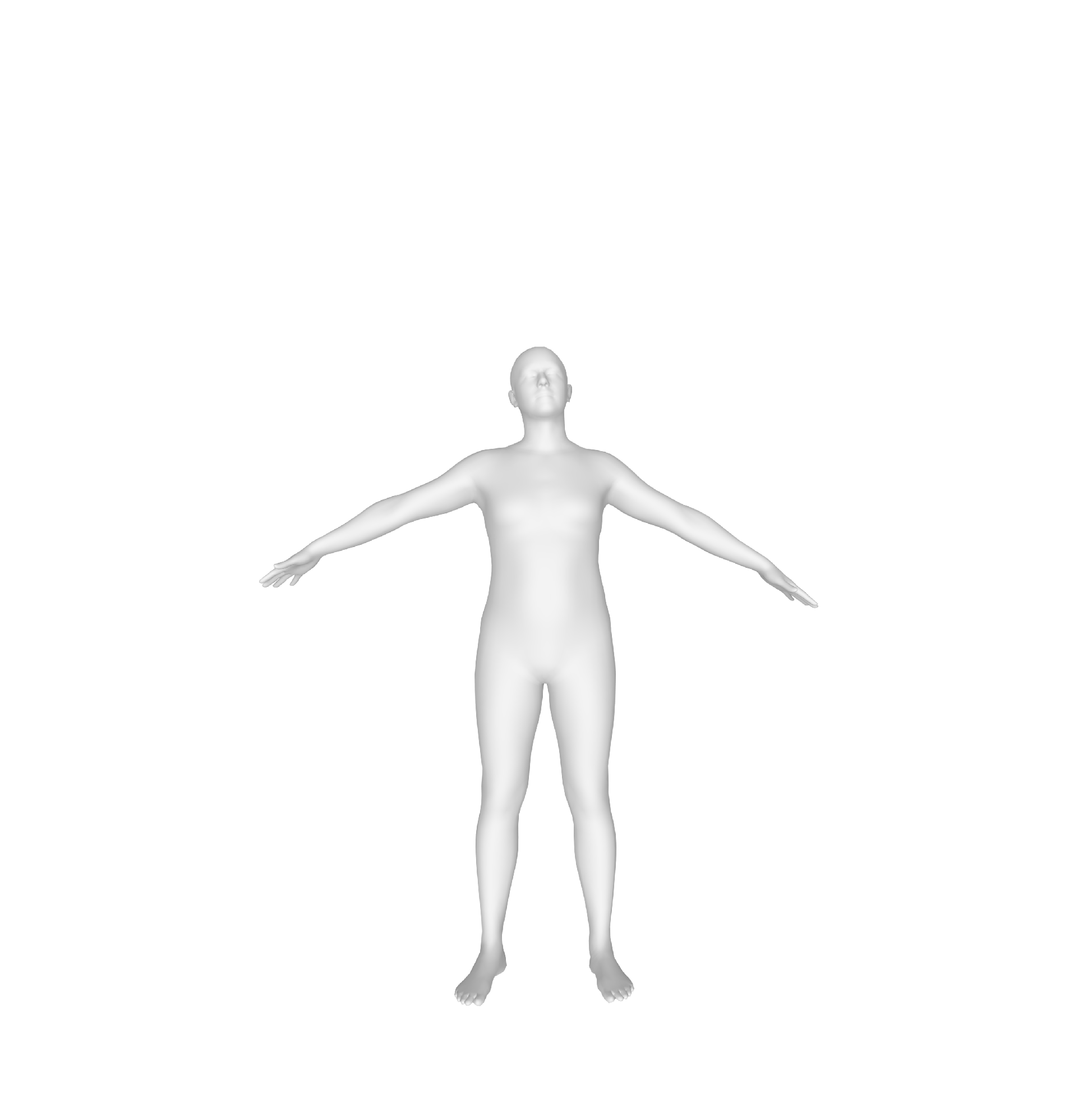}
        \caption{Ground Truth $t=500ms$}
    \end{subfigure}\hfill
    \begin{subfigure}{.185\linewidth}
        \includegraphics[trim=2.1cm 0.8cm 2.1cm 2.7cm, clip,width=\linewidth]{figures/modes_5_front.png}
        \caption{5 Modes $t=500ms$}
    \end{subfigure}
    
    \begin{subfigure}{.16\linewidth}
        \includegraphics[trim=2.1cm 0.8cm 2.1cm 2.7cm, clip, width=\linewidth]{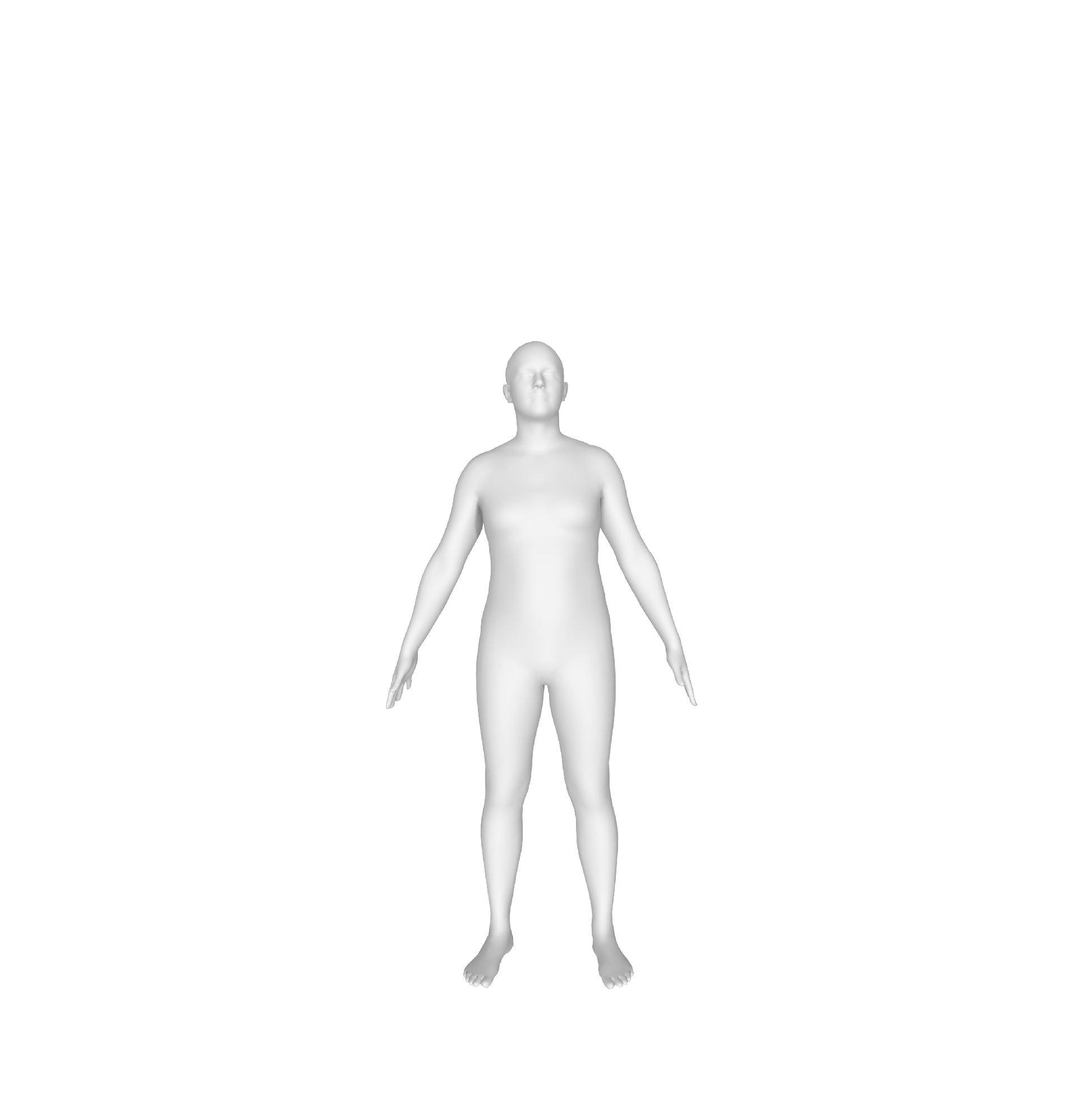}
        \caption{Ground Truth $t=1s$}
    \end{subfigure}{\unskip\ \vrule }
    \begin{subfigure}{.16\linewidth}
        \includegraphics[trim=2.1cm 0.8cm 2.1cm 2.7cm, clip, width=\linewidth]{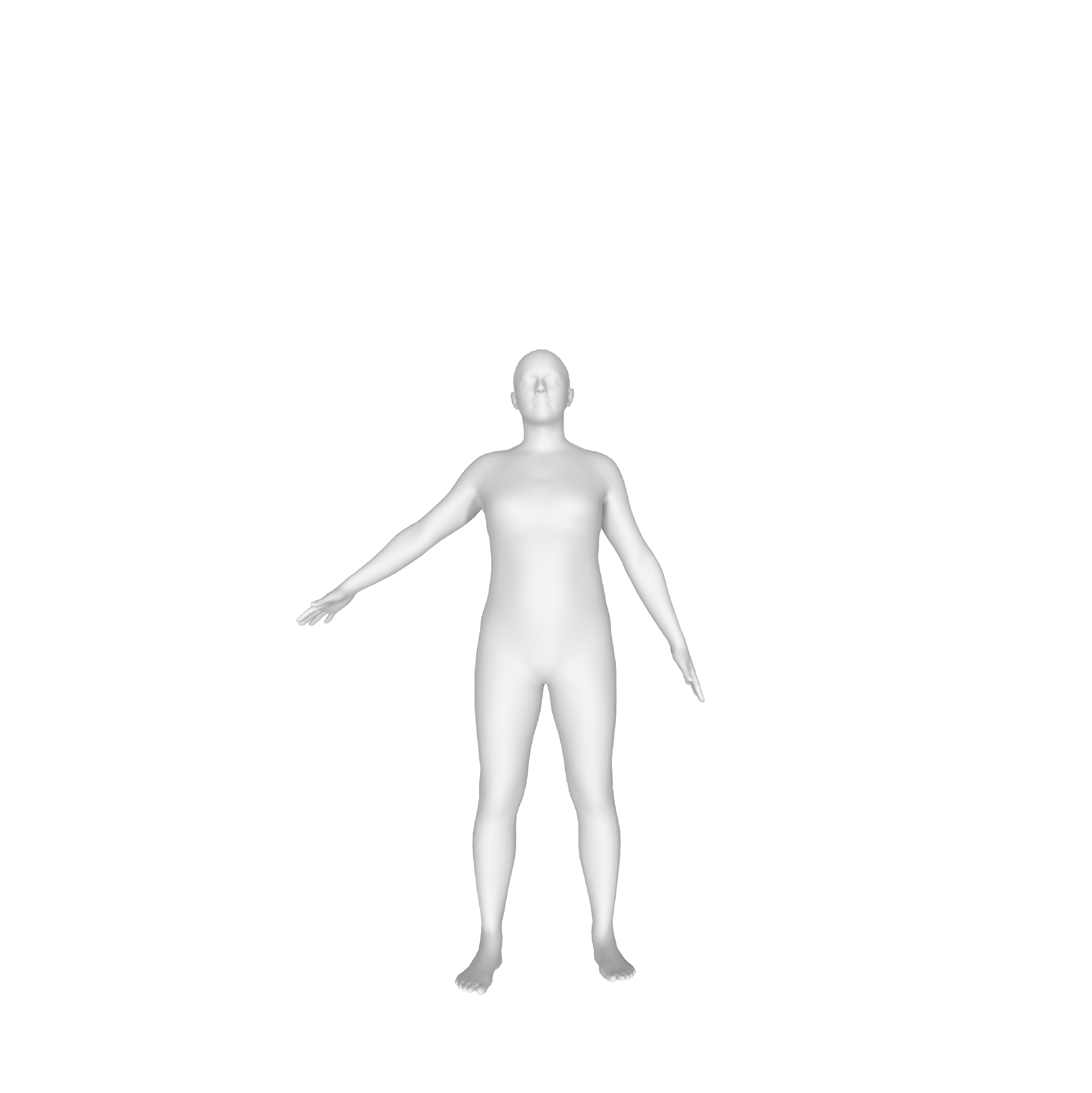}
        \caption{Mode 1 $p(z)=38\%$}
    \end{subfigure}
    \begin{subfigure}{.16\linewidth}
        \includegraphics[trim=2.1cm 0.8cm 2.1cm 2.7cm, clip, width=\linewidth]{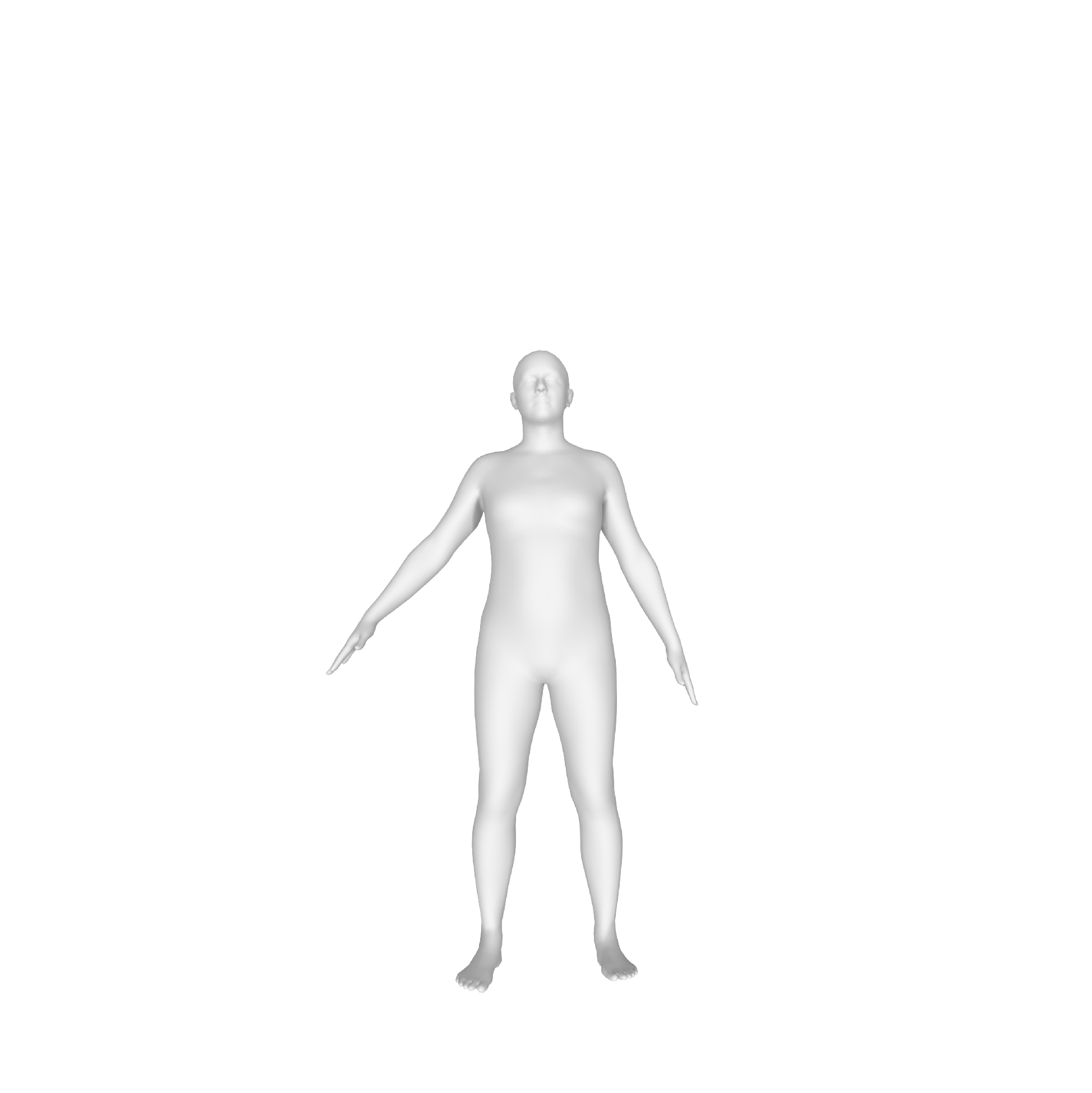}
        \caption{Mode 2 $p(z)=35\%$}
    \end{subfigure}
    \begin{subfigure}{.16\linewidth}
        \includegraphics[trim=2.1cm 0.8cm 2.1cm 2.7cm, clip, width=\linewidth]{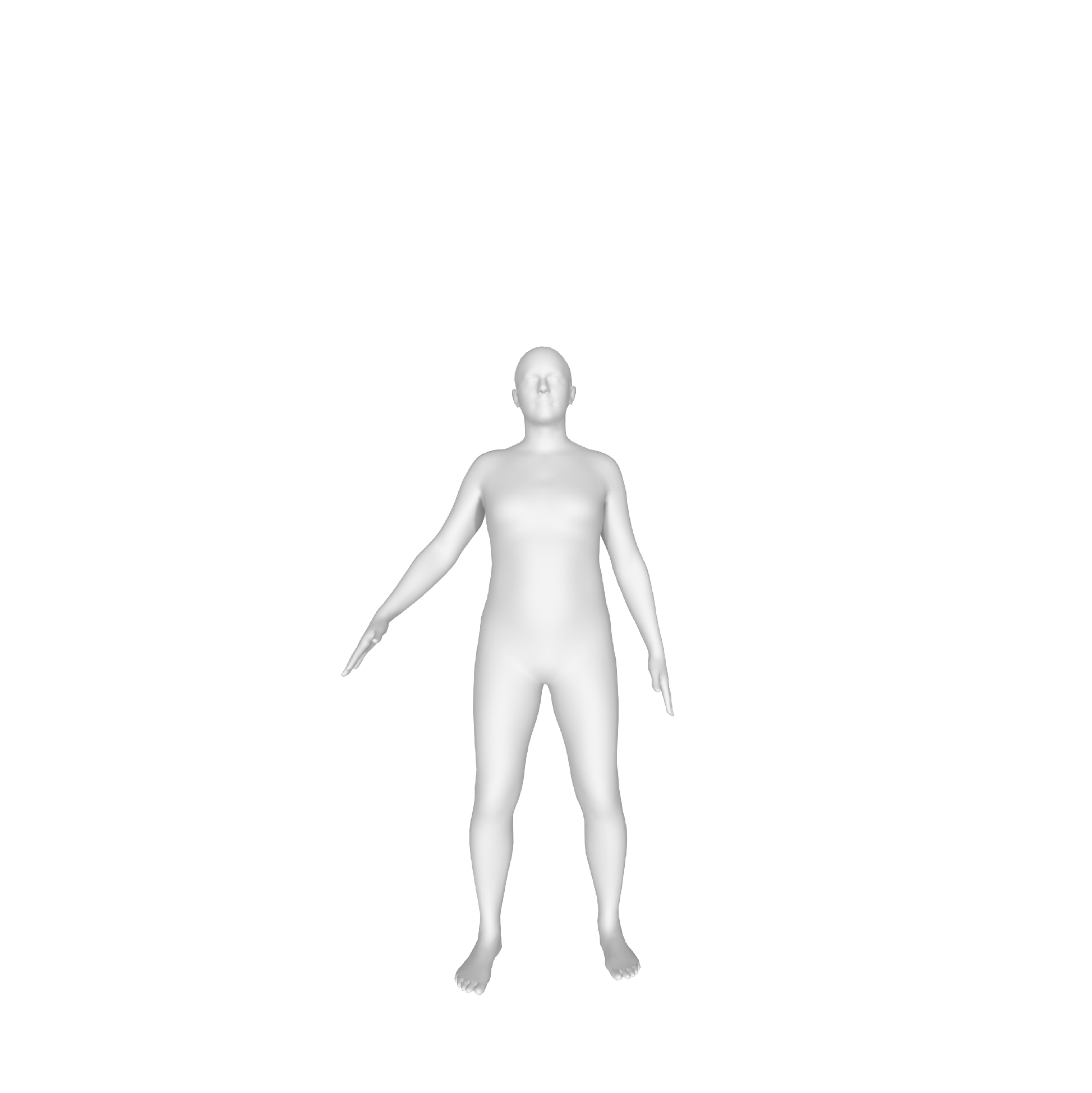}
        \caption{Mode 3 $p(z)=23\%$}
    \end{subfigure}
    \begin{subfigure}{.16\linewidth}
        \includegraphics[trim=2.1cm 0.8cm 2.1cm 2.7cm, clip, width=\linewidth]{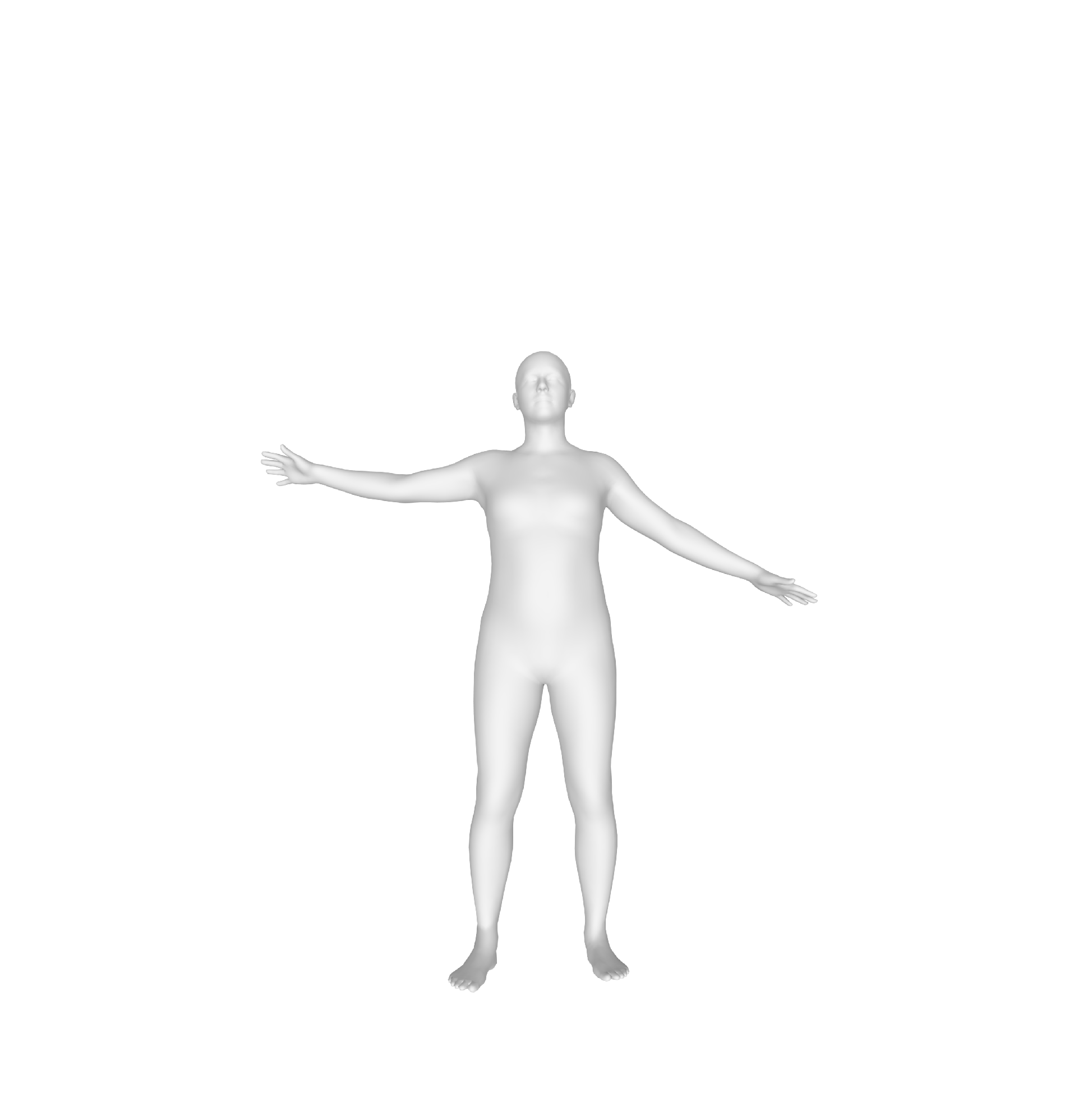}
        \caption{Mode 4 $p(z)=3\%$}
    \end{subfigure}
    \begin{subfigure}{.16\linewidth}
        \includegraphics[trim=2.1cm 0.8cm 2.1cm 2.7cm, clip, width=\linewidth]{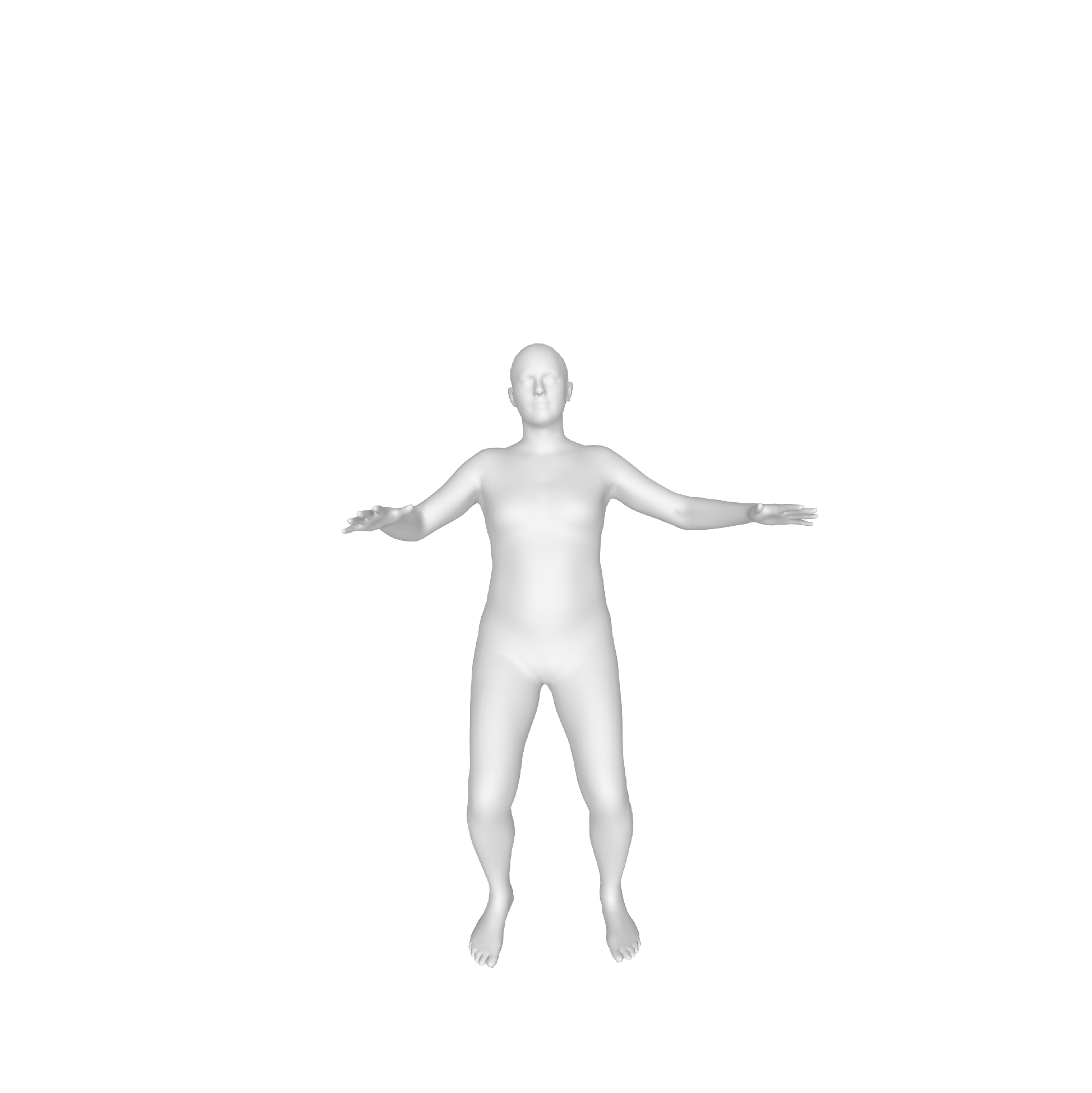}
        \caption{Mode 5 $p(z)=2\%$}
    \end{subfigure}
    
    \caption{Qualitative visualization of the prediction distribution on a single sample from the AMASS dataset. The human lands from a jump and pulls both arms downwards. (a)-(c), (e) Ground Truth poses. (d) 5 modes of our prediction distribution. Each mode is weighted by its probability where opacity indicates high confidence in a particular mode. (f)-(j) Mean of each of the five output modes at $t=1s$.}
    \label{fig:qualitative}
\end{figure*}
\subsection{Qualitative Results}\label{sec:qualitative}

\cref{fig:qualitative} shows the capability of our method to capture the multimodal nature of human motions. Displayed is an exemplary motion where the human lands from a jump and rapidly pulls his arms down. In the beginning $t=500ms$, the different latent modes capture different possible speeds of the downwards arm movement. While the most likely mode anticipates a slower downwards movement than performed, the second most likely mode captures the true motion closely. Further, less likely modes capture even faster motions as well as movements where the arms are pulled more in front of the body. %The legs and feet are predicted to become stationary, with low uncertainty, after the landing is performed. 
Notably, reasonably small uncertainty is presented by the model for other joints. Towards the end $t=1000ms$, the expressiveness and multimodality of the modes can be experienced as, for example, Mode 5 captures the possibility of a consecutive second jump.

\begin{figure}
    \centering
    \captionsetup[subfigure]{justification=centering}
    \begin{subfigure}{\linewidth}
        \begin{subfigure}{.24\linewidth}
            \includegraphics[trim=3.6cm 2.3cm 3.6cm 3.5cm, clip, width=\linewidth]{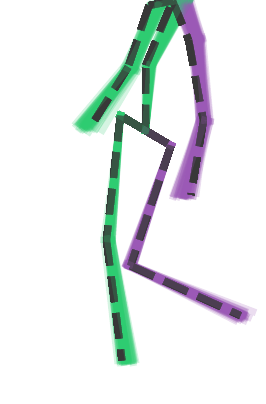}
            \caption*{No Occlusion \\ $t=40ms$}
        \end{subfigure}\vline
        \begin{subfigure}{.24\linewidth}
            \includegraphics[trim=3.6cm 2.3cm 3.6cm 3.5cm, clip, width=\linewidth]{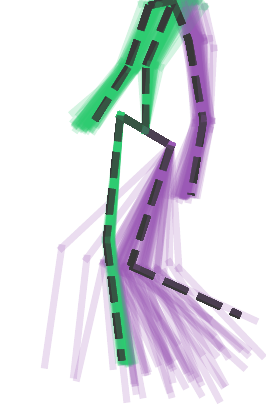}
            \caption*{Occlusion \\ $t=40ms$}
        \end{subfigure}\vline
        \begin{subfigure}{.24\linewidth}
            \includegraphics[trim=3.6cm 2.3cm 3.6cm 3.5cm, clip,width=\linewidth]{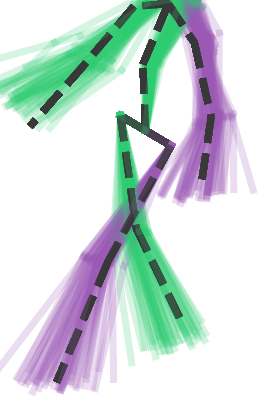}
            \caption*{No Occlusion \\ $t=400ms$}
        \end{subfigure}\vline
        \begin{subfigure}{.24\linewidth}
            \includegraphics[trim=3.6cm 2.3cm 3.6cm 3.5cm, clip,width=\linewidth]{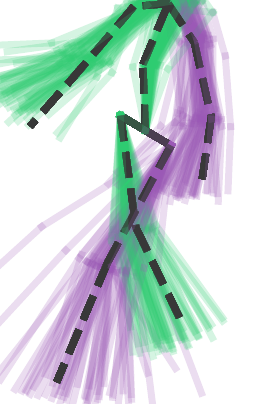}
            \caption*{Occlusion \\ $t=400ms$}
        \end{subfigure}
    \end{subfigure}
    \begin{subfigure}{0.9\linewidth}
        \includegraphics[width=\linewidth]{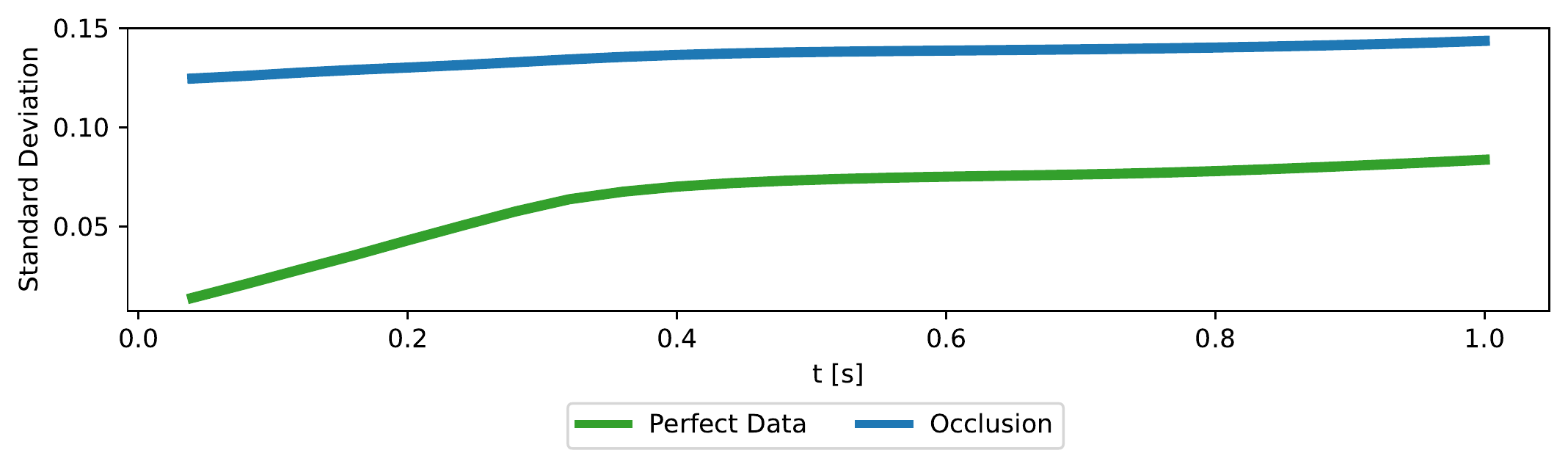}
    \end{subfigure}
    \caption{Handling of imperfect data. All joints of the left leg are artificially occluded. \textit{Top}: Visualization of prediction samples with and without occlusion. \textit{Bottom}: Mean standard deviation of parametric output distribution of left hip and knee. The occluded data is addressed by the model adjusting its own uncertainty. %\textit{only} for the joints with missing data (higher std). For later prediction timesteps the model gains relative confidence by taking in information from the other joints.
    \vspace{-0.4em}}
    \label{fig:occlusion}
\end{figure}

Another important quality for robotic applications is the ability to work with imperfect data. To simulate occlusions during training we apply \textit{Node Dropout}. For the occluded nodes, we set a random number of continuous states leading to $t=0$ to the neutral quaternion. The unique capabilities of our model here are shown in \cref{fig:occlusion}. In this instance, we artificially occlude all joint's data of the left leg. This leads to high variance but reasonable sampled predictions during early timesteps. More importantly, the model can understand and output its uncertainty: The closed form standard deviation of the parametric $\mathcal{N}^{\pi}_{SO(3)}$ output distribution is adequately higher compared to the distribution with perfect data. For increasing prediction time, the model uses the influence between nodes learned in the \textit{Typed Graph} layers to produce reasonable motions even for the occluded nodes and adjusts its relative confidence reasonably.

\vspace{-1.05em}
\section{Conclusion}
\vspace{-0.75em}
In this work, we present \emphalgname{}, a\rebuttalrm{generative} \rebuttal{probabilistic} human motion forecasting approach which uniquely provides the information plethora of a\rebuttalrm{generative} \rebuttal{probabilistic} approach and the accuracy of a deterministic model. Its predictions respect \rebuttal{rigid skeleton} constraints, all while producing full parametric motion distributions, which can be especially useful in downstream robotic applications. 
It achieves state-of-the-art prediction performance in a variety of metrics on standard and new real-world human motion datasets. Further, to the best of the authors' knowledge, it is the first method that demonstrates its ability to deal with occluded data while reasoning about its own uncertainty.

{\bf Limitations.} The approach is yet limited by the upstream data provider. While the motion capture system utilized to record the datasets used here, provides the required accuracy to calculate the joint rotations via inverse kinematics, the authors anticipate this being a challenge when relying on vision-based algorithms for human poses detection. \rebuttal{Further, our approach's tendency to less diverse predictions can become problematic with regards to new unseen behaviors where our approach would be overly confident.}

{\bf Future Directions} include incorporating \emphalgname{}'s human behavior predictions in downstream robotic planning, decision making, and control frameworks, as well as exploring options to tightly couple upstream vision algorithms.

%\section{Acknowledgement}

%%%%%%%%% REFERENCES
\printbibliography

\clearpage
\appendix
\section{Gradient Flow Through Latent Variable}\label{sec:GS_rep}
A reparameterization trick (Gumbel-Softmax) is commonly used to approximate the expectation $\mathbb{E}_{z \sim q_\psi(\cdot \mid x)}$ by taking $N$ samples from $q_\psi(z \mid x)$. We however, using the latent distribution in the output representation, exactly compute $\mathbb{E}_{z \sim q_\phi(\cdot \mid x)} = \sum_{z} \log q_\phi(z \mid x) p_\psi (y \mid x, z)$. 

Still the gradient resulting from each component's Log-Likelihood could be backpropagated to update $\phi$. Conceptually, we argue that the mixture distribution should not affect the individual component distributions’ log-prob computation. The same applies to the gradients. Practically, letting the gradient flow through the latent variable a marginal negative experimental influence on the performance. Thus, as the implementation difference is a single line, we will leave the choice to the user.

\section{Implementation and Training Details}\label{sec:details}
All model training was performed on a single \textit{Nvidia RTX2080}. 

\textbf{Deterministic Evaluation.} For the H3.6M dataset, we applied dataset augmentation (according to \cite{Pavllo2019ModelingNetworks}) where the training samples are mirrored along the human's vertical. A batch size of 32 and early stopping to select the best model was used. During hyperparameter optimization, we used subject 6 for validation purposes. For the final model, subject 6 was added to the training data, but we kept the same early stopping iteration which showed the best performance during validation. All test evaluations were performed on the full 32-joint skeleton of subject 5.
For the AMASS dataset we use 5\% of the training data samples as validation data. Again, we use early stopping based on the validation results. The batch size for AMASS is 128. All evaluations were performed on the full 22-joint skeleton.

\textbf{Generative Evaluation.} We use the same hyperparameters for the H3.6M dataset as in the deterministic evaluation. However, according to previous works, we only evaluate on the positions of a 16-joints skeleton where subject 9 and 11 serve as test subjects. To stabilize the GRU for the long prediction horizon (100 steps) we use a form of curriculum learning; increasing the prediction horizon from a small value to the target in early iterations.

For both model types we found it beneficial to randomly shrink the prediction horizon at every training iteration.

\newpage
\section{Multimodal Metrics}\label{sec:mm_metric}
\begin{figure}[ht]
    \centering
    \begin{subfigure}{\linewidth}
        \includegraphics[width=\linewidth]{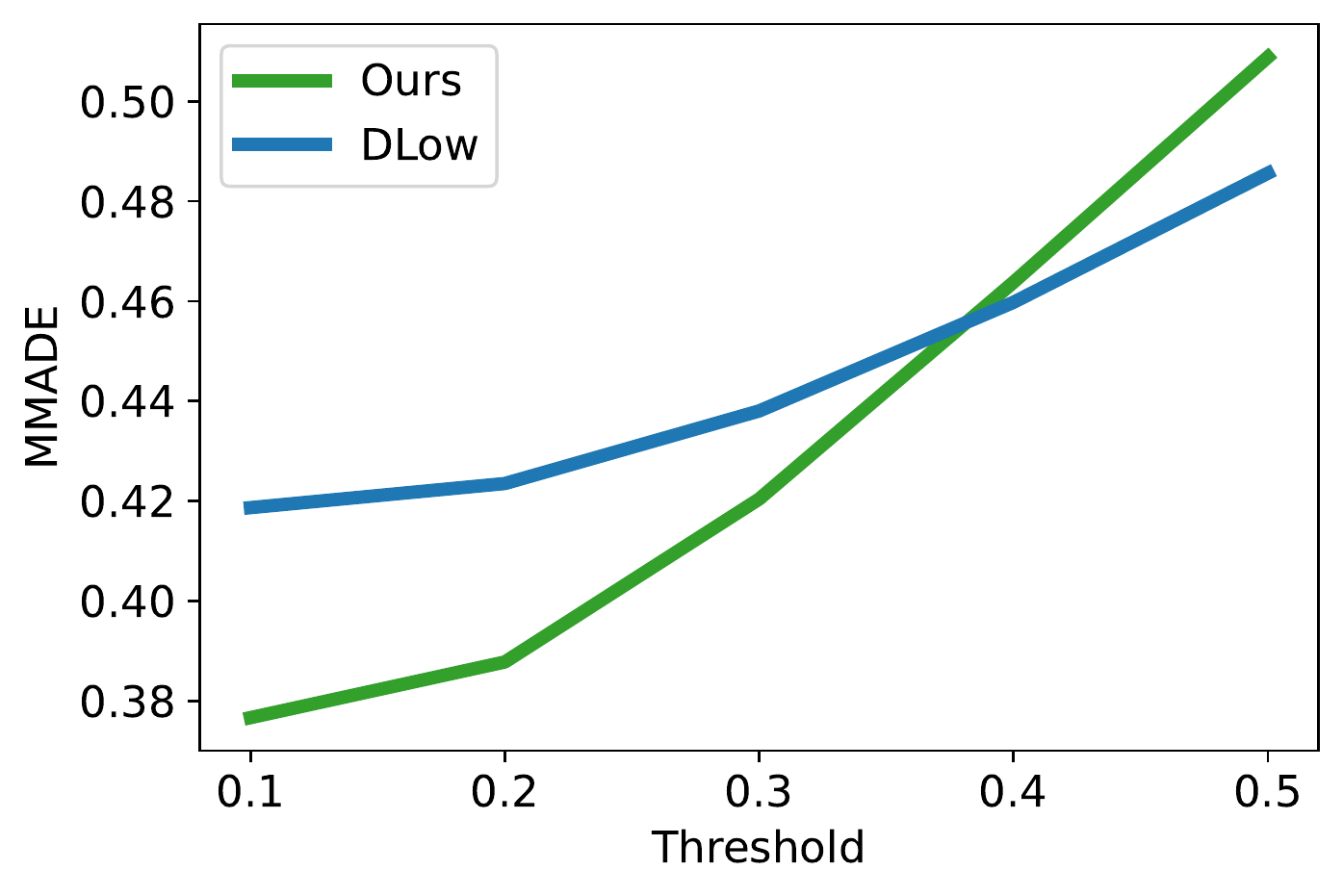}
        \caption{MMADE for different thresholds}
    \end{subfigure}
    \begin{subfigure}{\linewidth}
        \includegraphics[width=\linewidth]{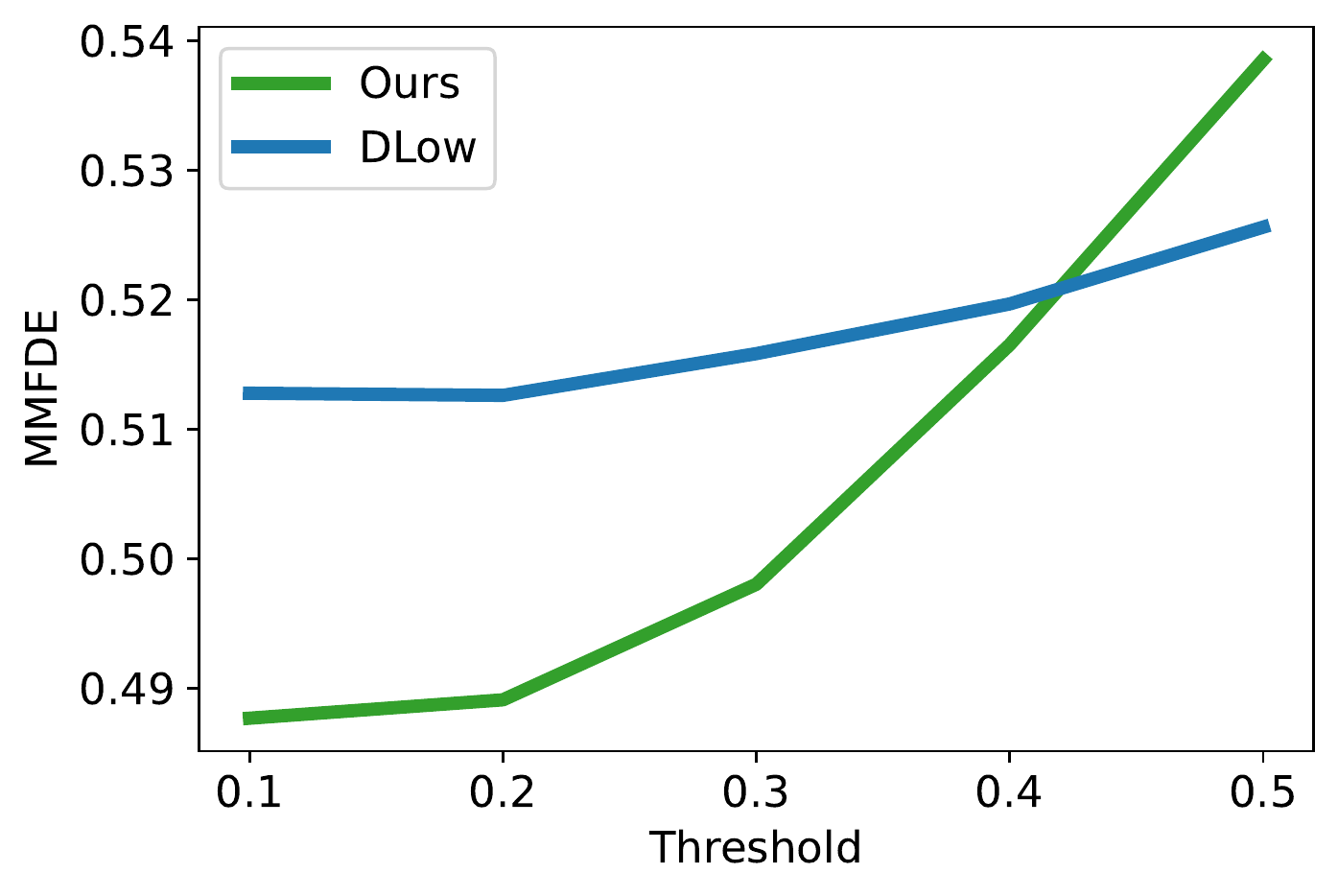}
        \caption{MMFDE for different thresholds}
    \end{subfigure}
    \caption{Result for multimodal metrics proposed in \cite{Yuan2020DiverseProcesses} for different thresholds. For larger thresholds, where possibly uncorrelated motions are evaluated together, \textit{DLow} achievs better numerical values as their approaches over-diversifies their produced motions.}
    \label{fig:mm_metric_plot}
\end{figure}

\begin{figure*}
    \centering
    \begin{subfigure}{0.225\linewidth}
        \includegraphics[width=\linewidth]{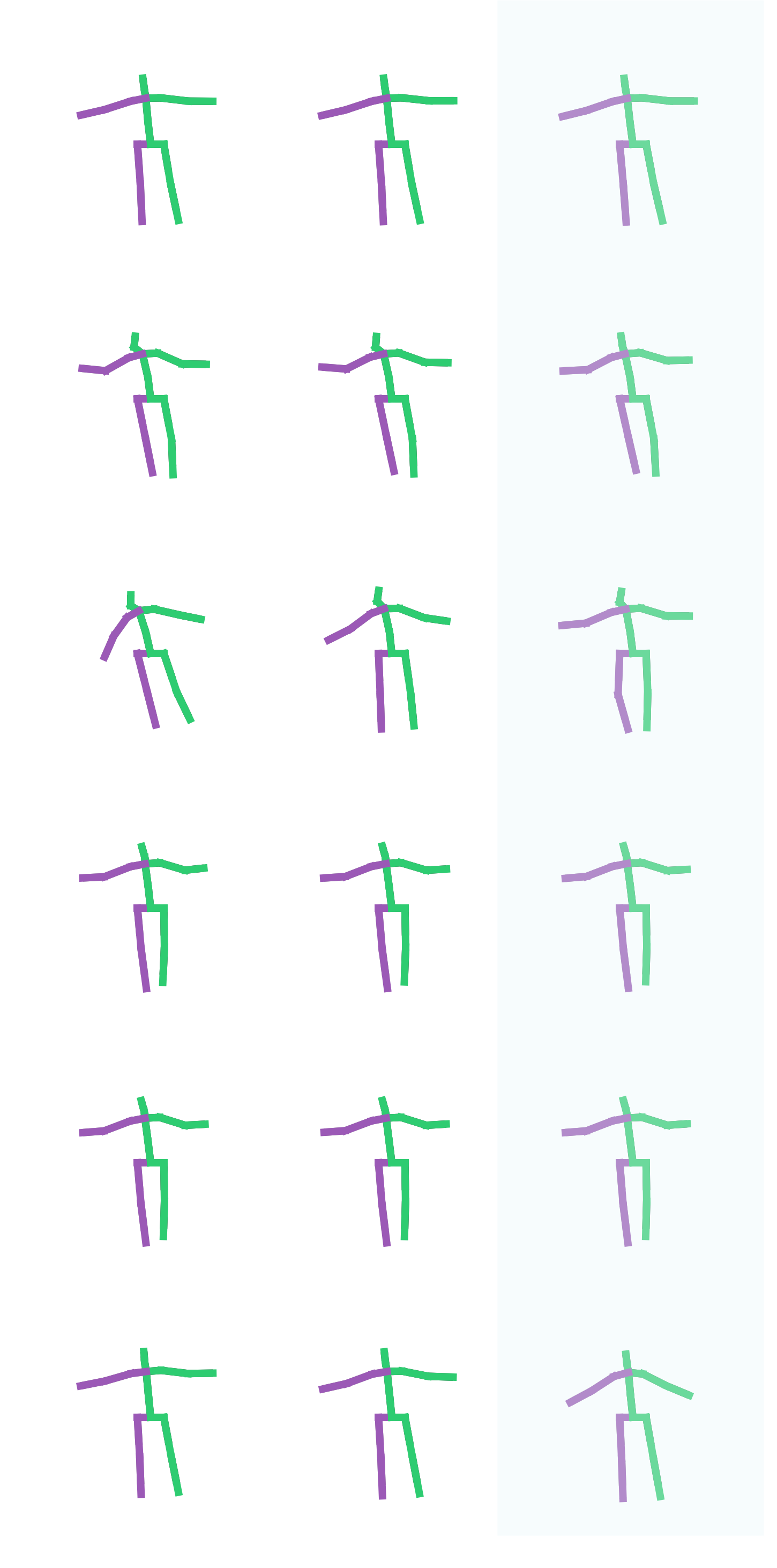}
        \caption*{$t: -500ms \hspace{0.5em} -250ms \hspace{2em} 0ms\hspace{1.5em}$}
    \end{subfigure}
    \vline
    \begin{subfigure}{0.765\linewidth}
        \includegraphics[width=\linewidth]{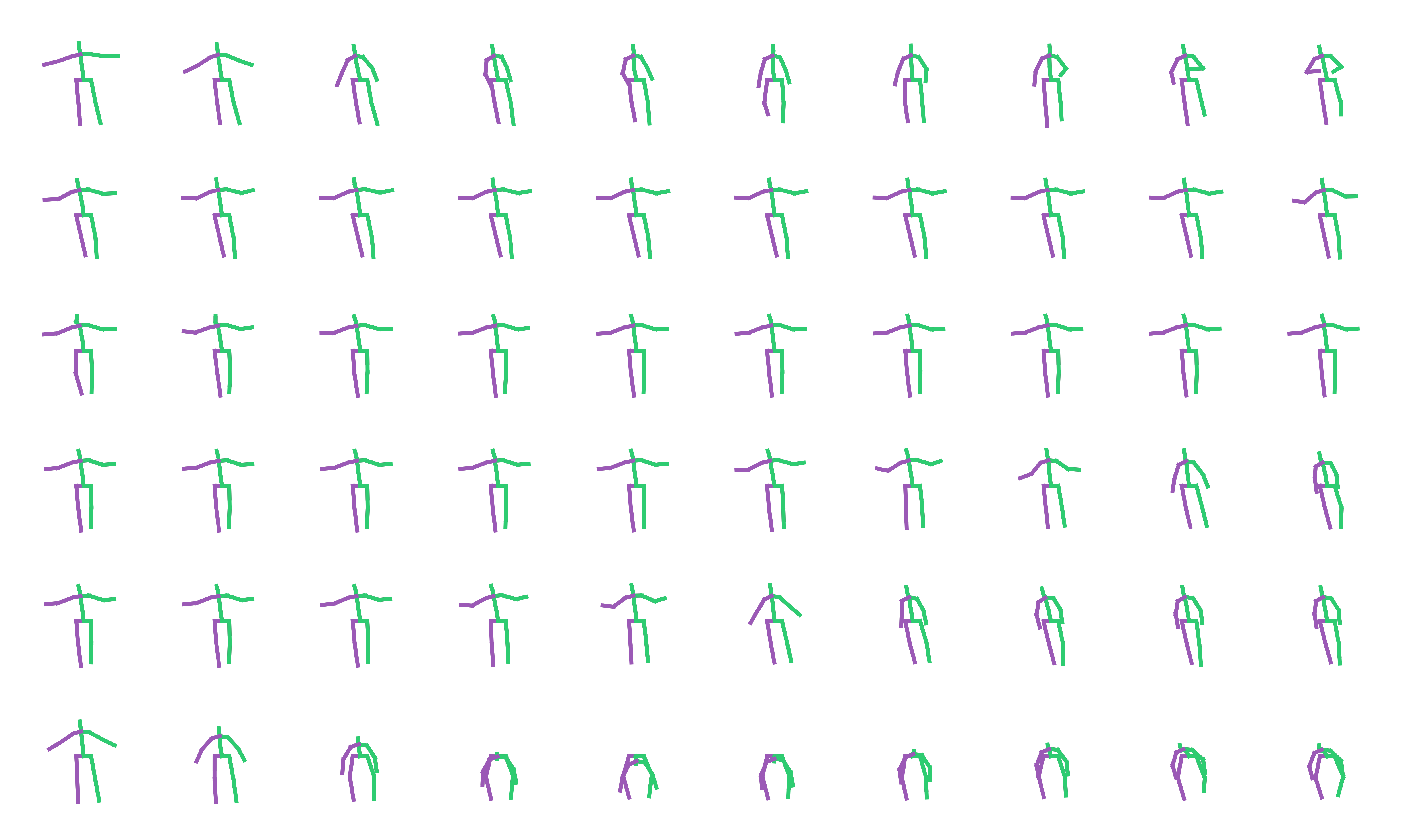}
        \caption*{$\hspace{1em} 200ms \hspace{2.0em} 400ms \hspace{2.0em} 600ms \hspace{2.0em} 800ms \hspace{2.0em} 1000ms \hspace{1.5em} 1200ms \hspace{1.5em} 1400ms \hspace{1.5em} 1600ms \hspace{1.5em} 1800ms \hspace{1.5em} 2000ms$}
    \end{subfigure}

    \caption{Example of 6 motions which are combined for the multimodal evaluation. At $t=0$ (highlighted) all motions seem to be similar. However, they greatly differ in history and future. Thus, we do not expect, and do not want, our model to give accurate output for all futures when only the history of the first motion is used as input.}
    \label{fig:mm_metric_ex}
\end{figure*}

In \cite{Yuan2020DiverseProcesses} the authors propose two multimodal evaluation metrics where motions are combined based on ''similar context'' and evaluate the mean performance on all futures of the combined motions. Their measure of similarity is solely based on the distance of motions at a single timestep $t=0$. Motions are combined for evaluation using a threshold on that distance. While this threshold is arbitrarily set to $0.5$ in \cite{Yuan2020DiverseProcesses}, we use a range of different thresholds in our evaluation. The results can be seen in \cref{fig:mm_metric_plot}. While we outperform \textit{DLow} for lower thresholds we have lower numerical results for higher thresholds. However, looking at a concrete example \cref{fig:mm_metric_ex} we see that for high thresholds motions with entirely uncorrelated futures (and history) are combined for evaluation. Thus, the worse numerical results for (too) high thresholds are expected; even encouraged here.

%\cref{fig:mm_metric} shows potential shortcoming of the multimodal metrics proposed in \cite{Yuan2019DiverseProcesses}. In their work, the authors argue for a multimodal metric by combining motions based on similar context and evaluate on all futures of the combined motions. However, their measure of similarity is solely based on the distance of motions at a single timestep. This can lead to motions being evaluated together which origin from entirely different high-level behaviors.

%\begin{figure}[h!]
%    \centering
%    \begin{subfigure}{\linewidth}
%        \fbox{\includegraphics[width=\linewidth]{figures/mm_metric.png}}
%        \caption{Original motivation figure for multimodal metric from \cite{Yuan2019DiverseProcesses}.}
%    \end{subfigure}
%    \begin{subfigure}{\linewidth}
%        \fbox{\includegraphics[width=\linewidth]{figures/mm_metric2.png}}
%        \caption{Adapted example to show failure.}
%    \end{subfigure}
%
%    \caption{In \cite{Yuan2019DiverseProcesses, Yuan2020DLow:Prediction} the introduction of multimodal metrics is motivated by Figure (a). The authors argue that by grouping motions with similar history (context) each future represents a mode of the ``same`` high-level motion. What is not clearly stated in \cite{Yuan2019DiverseProcesses}, but becomes clear looking at their publicly available implementation, is they use an arbitrary threshold at $t=0$ (dotted line) to define context similarity. Visualized by (b) this can lead to entirely different motions being grouped together.}
%    \label{fig:mm_metric}
%\end{figure}

\newpage
\section{Bone Deformation}\label{sec:bone_deform}
\begin{figure}[h!]
    \centering
    \includegraphics[width=\linewidth]{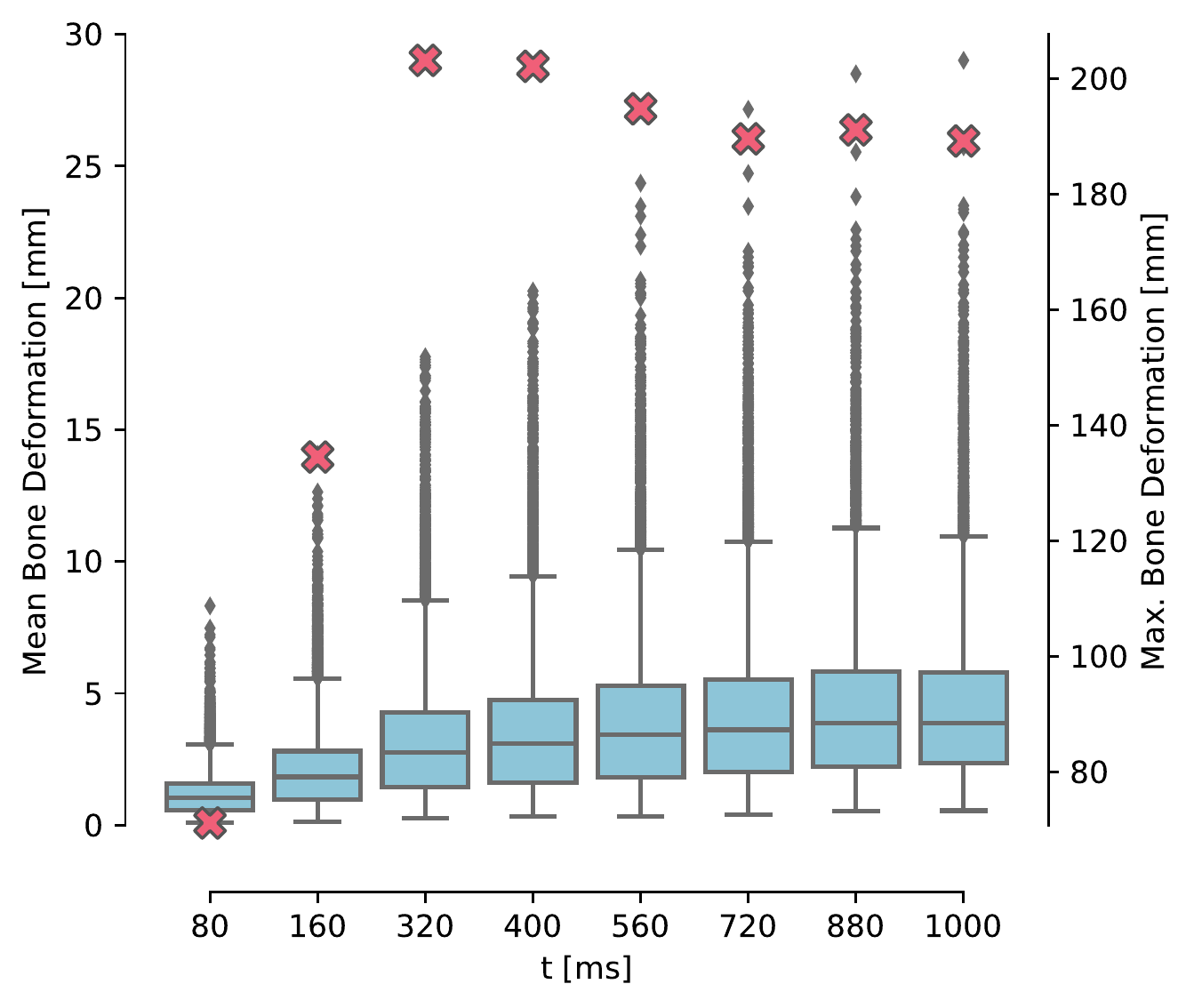}
    \caption{Exemplary bone deformation of a method directly predicting 3D joint positions. With increasing prediction time the deformation increases with outliers reaching deformations of over $20cm$. Boxplots: Mean bone deformation over all bones in the skeleton. Red Crosses: Maximum bone deformation over all bones. Analysis performed on ~10000 samples.}
    \label{fig:bone_deform}
\end{figure}

Some algorithms directly predict joint positions instead of joint configurations (angles). This shows advantages on metrics calculated on joint positions (e.g. MPJPE). However, as \cref{fig:bone_deform} outlines they produce physically unfeasible motions as the rigid bone structure is deformed.

\section{Learned Node-Attention}
\begin{figure*}
    \centering
    \includegraphics[width=\linewidth]{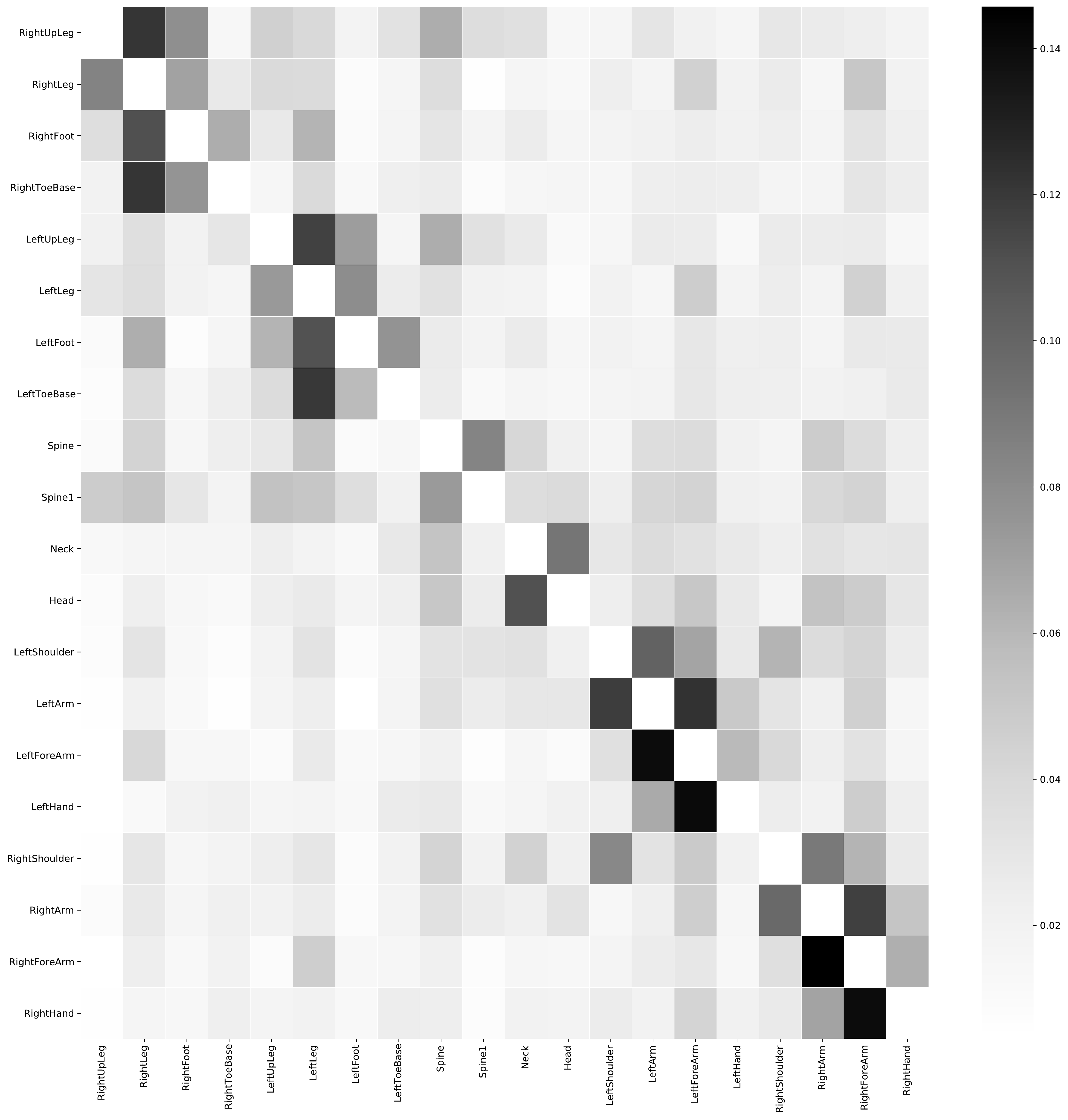}
    \caption{Learned attention influence of our TG layers between different nodes in the skeleton.}
    \label{fig:corr}
\end{figure*}

In \cref{fig:corr} we visualize the learned attention influence between different nodes in the skeleton by our \textit{Typed-Graph} approach. Unsurprisingly, neighboring joints commonly show high influence. However, influence between nodes not directly connecting by a single joint are learned too: The attention value connecting both shoulder joints has a high value as a correlation between the movements of both arms is learned. An example of a more subtle learned correlation is that the prediction of the hand is substantially influenced by the gaze angle of the head.

\section{Further Deterministic Results}\label{sec:add_det_res}
We present additional results for the deterministic evaluation. In \cref{tab:det_h36m_ang_ac} the results are split up by individual actions for 256 samples and for 8 samples in \cref{tab:det_h36m_ang_ac_8}. We also present the MPJPE metric which is calculated on the joint's position using forward kinematic on angle outputs (\cref{tab:det_h36m_mpjpe} and \cref{tab:det_amass_mpjpe}).

\begin{table*}[ht]
\footnotesize
\renewcommand\tabcolsep{2pt}
\begin{tabular*}{\linewidth}{@{\extracolsep{\fill}}lcccccccc|cccccccc|cccccccc@{}}
 & \multicolumn{8}{c}{Average} & \multicolumn{8}{c}{Walking} & \multicolumn{8}{c}{Eating} \\
milliseconds & 80 & 160 & 320 & 400 & 560 & 720 & 880 & 1000 & 80 & 160 & 320 & 400 & 560 & 720 & 880 & 1000 & 80 & 160 & 320 & 400 & 560 & 720 & 880 & 1000 \\ \hline
Zero Vel. & 0.40 & 0.70 & 1.11 & 1.25 & 1.46 & 1.63 & 1.76 & 1.84 & 0.43 & 0.78 & 1.23 & 1.34 & 1.50 & 1.56 & 1.58 & 1.55 & 0.28 & 0.53 & 0.89 & 1.03 & 1.18 & 1.33 & 1.44 & 1.49 \\
GRU sup. \cite{Martinez2017OnNetworks} & 0.43 & 0.74 & 1.15 & 1.30 & - & - & - & - & 0.34 & 0.60 & 0.91 & 0.98 & - & - & - & - & 0.30 & 0.57 & 0.87 & 0.98 & - & - & - & - \\
Quarternet \cite{Pavllo2019ModelingNetworks} & 0.37 & 0.62 & 1.00 & 1.14 & - & - & - & - & 0.28 & 0.49 & 0.76 & 0.83 & - & - & - & - & 0.22 & 0.47 & 0.76 & 0.88 & - & - & - & - \\
HistRepItself \cite{Mao2020HistoryAttention} & 0.28 & 0.52 & 0.88 & 1.02 & 1.23 & 1.40 & 1.55 & 1.64 & 0.24 & 0.43 & 0.66 & 0.71 & 0.84 & 0.91 & 0.99 & 1.03 & 0.18 & 0.37 & 0.67 & 0.79 & 0.95 & 1.11 & 1.23 & 1.30 \\
\hline
Ours W-Mean & 0.28 & 0.51 & 0.87 & 1.01 & 1.22 & 1.40 & 1.54 & 1.63 & 0.23 & 0.43 & 0.68 & 0.74 & 0.88 & 0.94 & 1.03 & 1.08 & 0.20 & 0.41 & 0.79 & 0.90 & 1.06 & 1.23 & 1.38 & 1.45 \\
Ours ML-Mode & 0.28 & 0.51 & 0.88 & 1.02 & 1.24 & 1.42 & 1.58 & 1.67 & 0.23 & 0.43 & 0.68 & 0.73 & 0.88 & 0.94 & 1.04 & 1.10 & 0.20 & 0.41 & 0.79 & 0.91 & 1.06 & 1.23 & 1.39 & 1.47 \\
\end{tabular*}

\begin{tabular*}{\linewidth}{@{\extracolsep{\fill}}lcccccccc|cccccccc|cccccccc@{}}
 & \multicolumn{8}{c}{Smoking} & \multicolumn{8}{c}{Discussion} & \multicolumn{8}{c}{Directions} \\
milliseconds & 80 & 160 & 320 & 400 & 560 & 720 & 880 & 1000 & 80 & 160 & 320 & 400 & 560 & 720 & 880 & 1000 & 80 & 160 & 320 & 400 & 560 & 720 & 880 & 1000 \\ \hline
Zero Vel. & 0.30 & 0.53 & 0.88 & 1.03 & 1.23 & 1.38 & 1.52 & 1.62 & 0.46 & 0.80 & 1.22 & 1.36 & 1.62 & 1.74 & 1.83 & 1.90 & 0.30 & 0.54 & 0.92 & 1.08 & 1.24 & 1.35 & 1.48 & 1.54 \\
GRU sup. \cite{Martinez2017OnNetworks} & 0.35 & 0.69 & 1.14 & 1.29 & - & - & - & - & 0.54 & 0.85 & 1.30 & 1.44 & - & - & - & - & 0.32 & 0.58 & 0.97 & 1.14 & - & - & - & - \\
Quarternet \cite{Pavllo2019ModelingNetworks} & 0.28 & 0.47 & 0.79 & 0.91 & - & - & - & - & 0.38 & 0.74 & 1.20 & 1.37 & - & - & - & - & 0.24 & 0.46 & 0.84 & 1.01 & - & - & - & - \\
HistRepItself \cite{Mao2020HistoryAttention} & 0.21 & 0.38 & 0.65 & 0.79 & 0.99 & 1.15 & 1.30 & 1.42 & 0.31 & 0.61 & 1.02 & 1.17 & 1.44 & 1.57 & 1.68 & 1.76 & 0.19 & 0.38 & 0.74 & 0.90 & 1.08 & 1.22 & 1.35 & 1.42 \\ 
\hline
Ours W-Mean & 0.21 & 0.38 & 0.65 & 0.80 & 1.02 & 1.32 & 1.34 & 1.44 & 0.36 & 0.63 & 1.02 & 1.16 & 1.41 & 1.54 & 1.63 & 1.70 & 0.19 & 0.38 & 0.75 & 0.94 & 1.13 & 1.26 & 1.38 & 1.46 \\
Ours ML-Mode & 0.21 & 0.38 & 0.65 & 0.80 & 1.03 & 1.22 & 1.36 & 1.47 & 0.35 & 0.63 & 1.02 & 1.18 & 1.44 & 1.58 & 1.68 & 1.75 & 0.19 & 0.38 & 0.76 & 0.95 & 1.15 & 1.28 &1.40 & 1.48
\end{tabular*}

\begin{tabular*}{\linewidth}{@{\extracolsep{\fill}}lcccccccc|cccccccc|cccccccc@{}}
 & \multicolumn{8}{c}{Greeting} & \multicolumn{8}{c}{Phoning} & \multicolumn{8}{c}{Posing} \\
milliseconds & 80 & 160 & 320 & 400 & 560 & 720 & 880 & 1000 & 80 & 160 & 320 & 400 & 560 & 720 & 880 & 1000 & 80 & 160 & 320 & 400 & 560 & 720 & 880 & 1000 \\ \hline
Zero Vel. & 0.54 & 0.91 & 1.41 & 1.59 & 1.81 & 2.01 & 2.16 & 2.23 & 0.39 & 0.69 & 1.14 & 1.28 & 1.48 & 1.70 & 1.85 & 1.91 & 0.40 & 0.74 & 1.20 & 1.40 & 1.72 & 2.00 & 2.29 & 2.44 \\
GRU sup. \cite{Martinez2017OnNetworks} & 0.64 & 0.99 & 1.40 & 1.54 & - & - & - & - & 0.42 & 0.70 & 1.11 & 1.27 & - & - & - & - & 0.46 & 0.83 & 1.33 & 1.52 & - & - & - & - \\
Quarternet \cite{Pavllo2019ModelingNetworks} & 0.61 & 0.93 & 1.34 & 1.51 & - & - & - & - & 0.36 & 0.61 & 0.98 & 1.14 & - & - & - & - & 0.38 & 0.71 & 1.20 & 1.39 & - & - & - & - \\
HistRepItself \cite{Mao2020HistoryAttention} & 0.39 & 0.71 & 1.17 & 1.35 & 1.60 & 1.78 & 1.93 & 1.99 & 0.29 & 0.52 & 0.91 & 1.05 & 1.24 & 1.47 & 1.63 & 1.72 & 0.27 & 0.55 & 1.00 & 1.21 & 1.54 & 1.80 & 2.10 & 2.24 \\ 
\hline
Ours W-Mean & 0.40 & 0.68 & 1.14 & 1.31 & 1.51 & 1.70 & 1.84 & 1.90 & 0.28 & 0.51 & 0.84 & 0.98 & 1.19 & 1.41 & 1.58 & 1.66 & 0.24 & 0.50 & 0.93 & 1.12 & 1.42 & 1.71 & 1.96 & 2.10 \\
Ours ML-Mode & 0.40 & 0.69 & 1.14 & 1.32 & 1.53 & 1.75 & 1.92 & 1.98 & 0.28 & 0.51 & 0.85 & 1.00 & 1.21 & 1.45 & 1.61 & 1.70 & 0.24 & 0.50 & 0.94 & 1.13 & 1.45 & 1.76 & 2.03 & 2.19 \\
\end{tabular*}

\begin{tabular*}{\linewidth}{@{\extracolsep{\fill}}lcccccccc|cccccccc|cccccccc@{}}
 & \multicolumn{8}{c}{Purchases} & \multicolumn{8}{c}{Sitting} & \multicolumn{8}{c}{Sitting Down} \\
milliseconds & 80 & 160 & 320 & 400 & 560 & 720 & 880 & 1000 & 80 & 160 & 320 & 400 & 560 & 720 & 880 & 1000 & 80 & 160 & 320 & 400 & 560 & 720 & 880 & 1000 \\ \hline
Zero Vel. & 0.57 & 0.96 & 1.36 & 1.44 & 1.64 & 1.79 & 1.94 & 2.01 & 0.33 & 0.60 & 1.01 & 1.16 & 1.41 & 1.67 & 1.87 & 1.98 & 0.50 & 0.84 & 1.30 & 1.48 & 1.99 & 2.21 & 2.31 \\
GRU sup. \cite{Martinez2017OnNetworks} & 0.57 & 0.95 & 1.33 & 1.43 & - & - & - & - & 0.41 & 0.75 & 1.22 & 1.41 & - & - & - & - & 0.59 & 1.00 & 1.62 & 1.87 & - & - & - & - \\
Quarternet \cite{Pavllo2019ModelingNetworks} & 054 & 0.92 & 1.36 & 1.47 & - & - & - & - & 0.34 & 0.59 & 1.00 & 1.15 & - & - & - & - & 0.47 & 0.81 & 1.31 & 1.50 & - & - & - & - \\
HistRepItself \cite{Mao2020HistoryAttention} & 0.43 & 0.78 & 1.21 & 1.31 & 1.47 & 1.62 & 1.74 & 1.81 & 0.25 & 0.49 & 0.91 & 1.06 & 1.33 & 1.59 & 1.78 & 1.88 & 0.41 & 0.72 & 1.17 & 1.36 & 1.66 & 1.88 & 2.11 & 2.20 \\ 
\hline
Ours W-Mean & 0.44 & 0.75 & 1.18 & 1.28 & 1.48 & 1.69 & 1.77 & 1.86 & 0.23 & 0.45 & 0.85 & 1.01 & 1.28 & 1.53 & 1.70 & 1.81 & 0.41 & 0.72 & 1.17 & 1.36 & 1.68 & 1.90 & 2.12 & 2.22 \\
Ours ML-Mode & 0.44 & 0.75 & 1.19 & 1.29 & 1.52 & 1.74 & 1.82 & 1.91 & 0.23 & 0.46 & 0.87 & 1.02 & 1.30 & 1.56 & 1.74 & 1.84 & 0.41 & 0.73 & 1.18 & 1.37 & 1.69 & 1.91 & 2.14 & 2.24 \\
\end{tabular*}

\begin{tabular*}{\linewidth}{@{\extracolsep{\fill}}lcccccccc|cccccccc|cccccccc@{}}
 & \multicolumn{8}{c}{Taking Photo} & \multicolumn{8}{c}{Waiting} & \multicolumn{8}{c}{Walk Dog} \\
milliseconds & 80 & 160 & 320 & 400 & 560 & 720 & 880 & 1000 & 80 & 160 & 320 & 400 & 560 & 720 & 880 & 1000 & 80 & 160 & 320 & 400 & 560 & 720 & 880 & 1000 \\ \hline
Zero Vel. & 0.26 & 0.44 & 0.73 & 0.84 & 1.05 & 1.20 & 1.33 & 1.45 & 0.37 & 0.64 & 1.10 & 1.27 & 1.50 & 1.67 & 1.91 & 1.92 & 0.53 & 0.85 & 1.20 & 1.31 & 1.46 & 1.63 & 1.73 & 1.78 \\
GRU sup. \cite{Martinez2017OnNetworks} & 0.30 & 0.52 & 0.88 & 1.02 & - & - & - & - & 0.41 & 0.68 & 1.20 & 1.37 & - & - & - & - & 0.52 & 0.84 & 1.21 & 1.32 & - & - & - & - \\
Quarternet \cite{Pavllo2019ModelingNetworks} & 0.23 & 0.39 & 0.69 & 0.81 & - & - & - & - & 0.32 & 0.54 & 1.00 & 1.15 & - & - & - & - & 0.48 & 0.78 & 1.12 & 1.21 & - & - & - & - \\
HistRepItself \cite{Mao2020HistoryAttention} & 0.19 & 0.34 & 0.60 & 0.72 & 0.92 & 1.07 & 1.21 & 1.33 & 0.25 & 0.46 & 0.88 & 1.05 & 1.28 & 1.47 & 1.63 & 1.75 & 0.41 & 0.68 & 1.01 & 1.12 & 1.30 & 1.45 & 1.54 & 1.63 \\ 
\hline
Ours W-Mean & 0.20 & 0.33 & 0.60 & 0.72 & 0.91 & 1.10 & 1.25 & 1.36 & 0.23 & 0.44 & 0.83 & 1.00 & 1.23 & 1.42 & 1.58 & 1.69 & 0.41 & 0.69 & 1.05 & 1.16 & 1.30 & 1.48 & 1.59 & 1.68 \\
Ours ML-Mode & 0.20 & 0.33 & 0.60 & 0.72 & 0.92 & 1.10 & 1.26 & 1.38 & 0.24 & 0.44 & 0.84 & 1.00 & 1.23 & 1.42 & 1.58 & 1.70 & 0.41 & 0.69 & 1.07 & 1.17 & 1.32 & 1.50 & 1.63 & 1.72 \\
\end{tabular*}

\begin{tabular}{lcccccccc}
 & \multicolumn{8}{c}{Walk Together} \\
milliseconds & 80 & 160 & 320 & 400 & 560 & 720 & 880 & 1000  \\ \hline
Zero Vel. & 0.37 & 0.66 & 1.02 & 1.15 & 1.32 & 1.39 & 1.41 & 1.43 \\
GRU sup. \cite{Martinez2017OnNetworks} & 0.35 & 0.57 & 0.83 & 0.94 & - & - & - & - \\
Quarternet \cite{Pavllo2019ModelingNetworks} & 0.28 & 0.45 & 0.69 & 0.79 & - & - & - & - \\
HistRepItself \cite{Mao2020HistoryAttention} & 0.21 & 0.38 & 0.62 & 0.71 & 0.86 & 0.94 & 1.00 & 1.04 \\ 
\hline
Ours W-Mean & 0.20 & 0.36 & 0.59 & 0.68 & 0.83 & 0.92 & 1.01 & 1.06 \\
Ours ML-Mode & 0.20 & 0.36 & 0.59 & 0.68 & 0.84 & 0.94 & 1.04 & 1.09 \\
\end{tabular}

\caption{Angle error on 256 samples per action on the H3.6M test dataset.}
\label{tab:det_h36m_ang_ac}
\end{table*}

\begin{table*}[ht]
\footnotesize
\renewcommand\tabcolsep{2pt}
\begin{tabular*}{\linewidth}{@{\extracolsep{\fill}}lcccccccc|cccccccc|cccccccc@{}}
 & \multicolumn{8}{c}{Average} & \multicolumn{8}{c}{Walking} & \multicolumn{8}{c}{Eating} \\
milliseconds & 80 & 160 & 320 & 400 & 560 & 720 & 880 & 1000 & 80 & 160 & 320 & 400 & 560 & 720 & 880 & 1000 & 80 & 160 & 320 & 400 & 560 & 720 & 880 & 1000 \\ \hline
Zero Vel. & 0.40 & 0.71 & 1.07 & 1.20 & 1.42 & 1.57 & 1.75 & 1.85 & 0.39 & 0.68 & 0.99 & 1.15 & 1.35 & 1.37 & 1.34 & 1.32 & 0.27 & 0.48 & 0.73 & 0.86 & 1.04 & 1.10 & 1.27 & 1.38 \\
GRU sup. \cite{Martinez2017OnNetworks} & 0.40 & 0.69 & 1.04 & 1.18 & - & - & - & - & 0.27 & 0.46 & 0.67 & 0.75 & 0.93 & - & - & 1.03 & 0.23 & 0.37 & 0.59 & 0.73 & 0.95 & - & - & 1.08  \\
DMGNN \cite{Li2020DynamicPrediction} & 0.27 & 0.52 & 0.83 & 0.95 & - & - & - & - & 0.18 & 0.31 & 0.49 & 0.58 & 0.66 & - & - & 0.75 & 0.17 & 0.30 & 0.49 & 0.59 & 0.74 & - & - & 1.14 \\
HistRepItself \cite{Mao2020HistoryAttention} & 0.27 & 0.52 & 0.82 & 0.93 & 1.14 & 1.28 & 1.48 & 1.59 & 0.18 & 0.30 & 0.46 & 0.51 & 0.59 & 0.62 & 0.61 & 0.64 & 0.16 & 0.29 & 0.49 & 0.60 & 0.74 & 0.81 & 1.01 & 1.11 \\
\hline
Ours W-Mean & 0.26 & 0.48 & 0.81 & 0.93 & 1.12 & 1.28 & 1.46 & 1.56 & 0.18 & 0.31 & 0.51 & 0.55 & 0.61 & 0.65 & 0.64 & 0.66 & 0.16 & 0.29 & 0.49 & 0.61 & 0.72 & 0.77 & 0.96 & 1.07 \\
Ours ML-Mode & 0.26 & 0.48 & 0.82 & 0.95 & 1.15 & 1.32 & 1.50 & 1.60 & 0.18 & 0.31 & 0.51 & 0.56 & 0.61 & 0.65 & 0.64 & 0.67 & 0.16 & 0.30 & 0.50 & 0.62 & 0.73 & 0.78 & 0.97 & 1.08\\
\end{tabular*}

\begin{tabular*}{\linewidth}{@{\extracolsep{\fill}}lcccccccc|cccccccc|cccccccc@{}}
 & \multicolumn{8}{c}{Smoking} & \multicolumn{8}{c}{Discussion} & \multicolumn{8}{c}{Directions} \\
milliseconds & 80 & 160 & 320 & 400 & 560 & 720 & 880 & 1000 & 80 & 160 & 320 & 400 & 560 & 720 & 880 & 1000 & 80 & 160 & 320 & 400 & 560 & 720 & 880 & 1000 \\ \hline
Zero Vel. & 0.26 & 0.48 & 0.97 & 0.95 & 1.02 & 1.14 & 1.47 & 1.69 & 0.31 & 0.67 & 0.94 & 1.04 & 1.41 & 1.71 & 1.86 & 1.96 & 0.39 & 0.59 & 0.79 & 0.89 & 1.02 & 1.22 & 1.47 & 1.50 \\
GRU sup. \cite{Martinez2017OnNetworks} & 0.32 & 0.59 & 1.01 & 1.10 & 1.25 & - & - & 1.50 & 0.30 & 0.67 & 0.98 & 1.06 & 1.43 & - & - & 1.69 & 0.41 & 0.64 & 0.80 & 0.92 & - & - & - & - \\
DMGNN \cite{Li2020DynamicPrediction} & 0.21 & 0.39 & 0.81 & 0.77 & 0.83 & - & - & 1.52 & 0.26 & 0.65 & 0.92 & 0.99 & 1.33 & - & - & 1.45 & 0.25 & 0.44 & 0.65 & 0.71 & - & - & - & - \\
HistRepItself \cite{Mao2020HistoryAttention} & 0.22 & 0.42 & 0.86 & 0.80 & 0.86 & 1.00 & 1.34 & 1.58 & 0.20 & 0.52 & 0.78 & 0.87 & 1.30 & 1.54 & 1.66 & 1.72 & 0.25 & 0.43 & 0.60 & 0.69 & 0.81 & 1.03 & 1.25 & 1.29 \\ 
\hline
Ours W-Mean & 0.21 & 0.42 & 0.79 & 0.89 & 0.96 & 1.02 & 1.33 & 1.48 & 0.21 & 0.56 & 0.80 & 0.90 & 1.23 & 1.51 & 1.57 & 1.53 & 0.29 & 0.38 & 0.63 & 0.72 & 0.87 & 1.08 & 1.29 & 1.34 \\
Ours ML-Mode & 0.21 & 0.43 & 0.82 & 0.93 & 1.01 & 1.09 & 1.40 & 1.56 & 0.21 & 0.56 & 0.81 & 0.92 & 1.26 & 1.54 & 1.61 & 1.58 & 0.29 & 0.39 & 0.67 & 0.77 & 0.97 & 1.16 & 1.37 & 1.42
\end{tabular*}

\begin{tabular*}{\linewidth}{@{\extracolsep{\fill}}lcccccccc|cccccccc|cccccccc@{}}
 & \multicolumn{8}{c}{Greeting} & \multicolumn{8}{c}{Phoning} & \multicolumn{8}{c}{Posing} \\
milliseconds & 80 & 160 & 320 & 400 & 560 & 720 & 880 & 1000 & 80 & 160 & 320 & 400 & 560 & 720 & 880 & 1000 & 80 & 160 & 320 & 400 & 560 & 720 & 880 & 1000 \\ \hline
Zero Vel. & 0.54 & 0.89 & 1.30 & 1.49 & 1.79 & 1.77 & 1.85 & 1.80 & 0.64 & 1.21 & 1.57 & 1.70 & 1.81 & 1.94 & 2.05 & 2.04 & 0.28 & 0.57 & 1.13 & 1.37 & 1.81 & 2.23 & 2.58 & 2.78 \\
GRU sup. \cite{Martinez2017OnNetworks} & 0.57 & 0.82 & 1.45 & 1.60 & - & - & - & - & 0.59 & 1.06 & 1.45 & 1.60 & - & - & - & - & 0.45 & 0.85 & 1.34 & 1.56 & - & - & - & - \\
DMGNN \cite{Li2020DynamicPrediction} & 0.36 & 0.61 & 0.94 & 1.12 & - & - & - & - & 0.52 & 0.97 & 1.29 & 1.43 & - & - & - & - & 0.20 & 0.46 & 1.06 & 1.34 & - & - & - & - \\
HistRepItself \cite{Mao2020HistoryAttention} & 0.35 & 0.60 & 0.95 & 1.14 & 1.48 & 1.47 & 1.61 & 1.57 & 0.53 & 1.01 & 1.22 & 1.29 & 1.42 & 1.55 & 1.68 & 1.68 & 0.19 & 0.46 & 1.09 & 1.35 & 1.59 & 1.83 & 2.14 & 2.34 \\ 
\hline
Ours W-Mean & 0.35 & 0.58 & 0.87 & 1.03 & 1.29 & 1.34 & 1.53 & 1.54 & 0.41 & 0.62 & 1.12 & 1.19 & 1.36 & 1.39 & 1.53 & 1.69 & 0.19 & 0.48 & 1.03 & 1.25 & 1.55 & 1.96 & 2.22 & 2.39 \\
Ours ML-Mode & 0.35 & 0.59 & 0.87 & 1.02 & 1.28 & 1.31 & 1.50 & 1.53 & 0.42 & 0.63 & 1.16 & 1.24 & 1.48 & 1.54 & 1.66 & 1.81 & 0.19 & 0.48 & 1.03 & 1.26 & 1.66 & 2.12 & 2.40 & 2.58 \\
\end{tabular*}

\begin{tabular*}{\linewidth}{@{\extracolsep{\fill}}lcccccccc|cccccccc|cccccccc@{}}
 & \multicolumn{8}{c}{Purchases} & \multicolumn{8}{c}{Sitting} & \multicolumn{8}{c}{Sitting Down} \\
milliseconds & 80 & 160 & 320 & 400 & 560 & 720 & 880 & 1000 & 80 & 160 & 320 & 400 & 560 & 720 & 880 & 1000 & 80 & 160 & 320 & 400 & 560 & 720 & 880 & 1000 \\ \hline
Zero Vel. & 0.62 & 0.88 & 1.19 & 1.27 & 1.64 & 1.62 & 2.09 & 2.45 & 0.40 & 0.63 & 1.02 & 1.18 & 1.26 & 1.36 & 1.57 & 1.63 & 0.39 & 0.74 & 1.07 & 1.19 & 1.36 & 1.57 & 1.70 & 1.80 \\
GRU sup. \cite{Martinez2017OnNetworks} & 0.58 & 0.79 & 1.08 & 1.15 & - & - & - & - & 0.41 & 0.68 & 1.12 & 1.33 & - & - & - & - & 0.47 & 0.88 & 1.37 & 1.54 & - & - & - & - \\
DMGNN \cite{Li2020DynamicPrediction} & 0.41 & 0.61 & 1.05 & 1.14 & - & - & - & - & 0.26 & 0.42 & 0.76 & 0.97 & - & - & - & - & 0.32 & 0.65 & 0.93 & 1.05 & - & - & - & - \\
HistRepItself \cite{Mao2020HistoryAttention} & 0.42 & 0.65 & 1.00 & 1.07 & 1.43 & 1.53 & 1.94 & 2.24 & 0.29 & 0.47 & 0.83 & 1.01 & 1.16 & 1.29 & 1.50 & 1.55 & 0.30 & 0.63 & 0.92 & 1.04 & 1.18 & 1.42 & 1.55 & 1.69 \\ 
\hline
Ours W-Mean & 0.41 & 0.64 & 1.06 & 1.14 & 1.42 & 1.66 & 2.00 & 2.28 & 0.26 & 0.44 & 0.79 & 0.99 & 1.12 & 1.28 & 1.52 & 1.58 & 0.29 & 0.60 & 0.93 & 1.08 & 1.30 & 1.51 & 1.64 & 1.78 \\
Ours ML-Mode & 0.42 & 0.64 & 1.07 & 1.17 & 1.44 & 1.73 & 2.08 & 2.33 & 0.26 & 0.43 & 0.78 & 0.98 & 1.09 & 1.25 & 1.49 & 1.55 & 0.29 & 0.60 & 0.94 & 1.10 & 1.30 & 1.50 & 1.63 & 1.76 \\
\end{tabular*}

\begin{tabular*}{\linewidth}{@{\extracolsep{\fill}}lcccccccc|cccccccc|cccccccc@{}}
 & \multicolumn{8}{c}{Taking Photo} & \multicolumn{8}{c}{Waiting} & \multicolumn{8}{c}{Walk Dog} \\
milliseconds & 80 & 160 & 320 & 400 & 560 & 720 & 880 & 1000 & 80 & 160 & 320 & 400 & 560 & 720 & 880 & 1000 & 80 & 160 & 320 & 400 & 560 & 720 & 880 & 1000 \\ \hline
Zero Vel. & 0.25 & 0.51 & 0.79 & 0.92 & 1.03 & 1.13 & 1.22 & 1.27 & 0.34 & 0.67 & 1.22 & 1.47 & 1.89 & 2.27 & 2.57 & 2.63 & 0.60 & 0.98 & 1.36 & 1.50 & 1.74 & 1.87 & 1.95 & 1.96 \\
GRU sup. \cite{Martinez2017OnNetworks} & 0.28 & 0.57 & 0.90 & 1.02 & - & - & - & - & 0.32 & 0.63 & 1.07 & 1.26 & - & - & - & - & 0.52 & 0.89 & 1.25 & 1.40 & - & - & - & - \\
DMGNN \cite{Li2020DynamicPrediction} & 0.15 & 0.34 & 0.58 & 0.71 & - & - & - & - & 0.22 & 0.49 & 0.88 & 1.10 & - & - & - & - & 0.42 & 0.72 & 1.16 & 1.34 & - & - & - & - \\
HistRepItself \cite{Mao2020HistoryAttention} & 0.16 & 0.36 & 0.58 & 0.70 & 0.83 & 0.91 & 1.01 & 1.08 & 0.22 & 0.49 & 0.92 & 1.14 & 1.54 & 1.92 & 2.25 & 2.33 & 0.46 & 0.78 & 1.05 & 1.23 & 1.58 & 1.65 & 1.79 & 1.85  \\ 
\hline
Ours W-Mean & 0.13 & 0.34 & 0.58 & 0.70 & 0.78 & 0.83 & 0.89 & 0.97 & 0.20 & 0.46 & 0.88 & 1.09 & 1.46 & 1.77 & 2.05 & 2.11 & 0.41 & 0.71 & 1.15 & 1.32 & 1.52 & 1.69 & 1.77 & 1.79 \\
Ours ML-Mode & 0.14 & 0.34 & 0.57 & 0.70 & 0.79 & 0.84 & 0.88 & 0.96 & 0.20 & 0.45 & 0.87 & 1.08 & 1.46 & 1.77 & 2.06 & 2.12 & 0.41 & 0.72 & 1.16 & 1.33 & 1.53 & 1.71 & 1.79 & 1.83 \\
\end{tabular*}

\begin{tabular}{lcccccccc}
 & \multicolumn{8}{c}{Walk Together} \\
milliseconds & 80 & 160 & 320 & 400 & 560 & 720 & 880 & 1000  \\ \hline
Zero Vel. & 0.33 & 0.66 & 0.94 & 0.99 & 1.10 & 1.22 & 1.22 & 1.52 \\
GRU sup. \cite{Martinez2017OnNetworks} & 0.27 & 0.53 & 0.74 & 0.79 & - & - & - & - \\
DMGNN \cite{Li2020DynamicPrediction} & 0.15 & 0.33 & 0.50 & 0.57 & - & - & - & - \\
HistRepItself \cite{Mao2020HistoryAttention} & 0.14 & 0.32 & 0.50 & 0.55 & 0.63 & 0.68 & 0.80 & 1.18 \\ 
\hline
Ours W-Mean & 0.13 & 0.31 & 0.48 & 0.54 & 0.67 & 0.78 & 0.92 & 1.20 \\
Ours ML-Mode & 0.13 & 0.31 & 0.49 & 0.55 & 0.70 & 0.81 & 0.98 & 1.27 \\
\end{tabular}

\caption{Angle error on 8 samples per action on the H3.6M test dataset.}
\label{tab:det_h36m_ang_ac_8}
\end{table*}

\clearpage
\begin{table*}[ht]
\footnotesize
\renewcommand\tabcolsep{3pt}
\begin{tabular*}{\linewidth}{@{\extracolsep{\fill}}lcccccccc@{}}
 & \multicolumn{8}{c}{MPJPE [mm]} \\
milliseconds & 80 & 160 & 320 & 400 & 560 & 720 & 880 & 1000 \\ \hline
Zero Vel. &  23.8 & 44.4 & 76.1 & 88.3 & 107.5 & 121.6 & 131.6 & 136.6 \\
HistRepItself \cite{Mao2020HistoryAttention} &  12.4 & 25.9 & 51.4 & 62.5 & 81.4 & 96.3 & 108.6 & 116.4 \\
HistRepItself 3D & 11.3 & 24.1 & 49.9 & 60.8 & 78.3 & 92.0 & 105.1 & 112.8 \\ 
\hline
Ours W-Mean & 15.1 & 29.9 & 55.6 & 66.2 & 83.7 & 98.0 & 110.1 & 117.9 \\
Ours ML-Mode & 15.0 & 30.1 & 56.0 & 66.6 & 84.3 & 98.9 & 111.4 & 119.6 \\
\hline
Ours Bo3-Modes & 15.1 & 29.6 & 53.6 & 63.1 & 78.4 & 91.0 & 102.4 & 110.5 \\
Ours Bo5-Modes & 15.2 & 29.7 & 53.2 & 62.6 & 78.4 & 88.2 & 99.3 & 107.7 \\
\end{tabular*}
\caption{Mean per Joint Position Error (MPJPE) on 256 samples per action on the H3.6M test dataset. HistRepItself 3D directly outputs 3D joint position and is therefore subject to bone deformation.}
\label{tab:det_h36m_mpjpe}
\end{table*}

\begin{table*}
\footnotesize
\renewcommand\tabcolsep{4pt}
\begin{tabular*}{\linewidth}{@{\extracolsep{\fill}}lcccccc@{}}
 & \multicolumn{6}{c}{MPJPE [mm]} \\
milliseconds & 100 & 200 & 400 & 600 & 800 & 1000 \\ \hline
Zero Vel. & 41.9 & 72.9 & 106.2 & 115.3 & 115.3 & 112.5 \\
HistRepItself \cite{Mao2020HistoryAttention} & 20.4 & 39.8 & 64.4 & 74.8 & 80.5 & 85.8 \\
\hline
Ours W-Mean & 19.1 & 37.8 & 63.0 & 75.3 & 82.3 & 87.5\\
Ours ML-Mode & 19.1 & 38.2 & 64.1 & 76.9 & 84.3 & 89.9 \\
\hline
Ours Bo3-Modes & 19.0 & 37.3 & 59.9 & 70.0 & 76.6 & 82.8 \\
Ours Bo5-Modes & 19.1 & 37.4 & 58.8 & 67.5 & 73.8 & 81.0
\end{tabular*}
\caption{Mean per Joint Position Error (MPJPE) on 10,000 samples from the AMASS test set.}
\label{tab:det_amass_mpjpe}
\end{table*}

\end{document}